# Image space formalism of convolutional neural networks for k-space interpolation

P. Dawood[1,2*], F. Breuer[3], M. Gram[1,4], I. Homolya[5], P.M. Jakob[1], M. Zaiss[2] and M. Blaimer[3]

[1] Experimental Physics 5, University of Würzburg, Würzburg, Germany

[2] Institute of Neuroradiology, University Hospital Erlangen, Erlangen, Germany

[3] Magnetic Resonance and X-ray Imaging Department, Fraunhofer Institute for Integrated Circuits IIS, Division Development Center X-Ray Technology, Würzburg, Germany

[4]Department of Internal Medicine I, University Hospital Würzburg, Würzburg, Germany

[5] Molecular and Cellular Imaging, Comprehensive Heart Failure Center, University Hospital Würzburg, Würzburg, Germany

*Correspondence to: Peter Dawood

Department of Physics, University of Würzburg

Experimental Physics 5

Am Hubland, 97074 Würzburg, Germany

Email: peter.dawood@uni-wuerzburg.de

## Abstract

**Purpose** Noise resilience in image reconstructions by scan-specific robust artificial neural networks for k-space interpolation (RAKI) is linked to nonlinear activations in k-space. To gain a deeper understanding of this relationship, an image space formalism of RAKI is introduced for analyzing noise propagation analytically, identifying and characterizing image reconstruction features and to describe the role of nonlinear activations in a human readable manner.

**Methods** The image space formalism for RAKI inference is employed by expressing nonlinear activations in k-space as element-wise multiplications with activation masks, which transform into convolutions in image space. Jacobians of the de-aliased, coil-combined image relative to the aliased coil images can be expressed algebraically, and thus, the noise amplification is quantified analytically (g-factor maps). We analyze the role of nonlinearity for noise resilience by controlling the degree of nonlinearity in the reconstruction model via the negative slope parameter in leaky ReLU.

**Results** The analytical g-factor maps correspond with those obtained from Monte Carlo simulations and from an auto differentiation approach for in vivo brain images. Apparent blurring and contrast loss artifacts are identified as implications of enhanced noise resilience. These residual artifacts can be traded against noise resilience by adjusting the degree of nonlinearity in the model (Tikhonov-like regularization) in case of limited training data. The inspection of image space activations reveals an autocorrelation pattern leading to a potential center artifact.

**Conclusion** The image space formalism of RAKI provides the means for analytical quantitative noise-propagation analysis and human-readable visualization of the effects of the nonlinear activation functions in k-space.

Keywords: RAKI, GRAPPA, Parallel Imaging, convolutional neural network, g-factor, noise propagation, explainable AI

## 1. Introduction

Gradient subsampling is the standard way to reduce the scan time in MRI. To obtain artifact-free images from the subsampled data, parallel imaging (PI) algorithms have been developed. For example, GRAPPA (1) is a widely used method operating in k-space. It interpolates missing k-space signals by convolution of acquired multichannel signals. The convolution filters are typically calibrated using a small amount of additionally fully-sampled data in the k-space center (auto-calibration signals, ACS), in which the signal-to-noise ratio (SNR) is higher. However, at high undersampling rates, the matrix systems are ill-conditioned and introduce severe spatially varying noise amplification in the reconstructed (i.e., de-aliased and coil-combined) images. Over the past decades, the basic principle of GRAPPA has been extended (2). For example, SPiRIT(3) allows k-space interpolation of non-uniformly sampled or non-Cartesian acquisitions and enforces consistency between acquired and estimated k-space data iteratively. Another k-space interpolation technique relying on the annihilation principle is AC-LORAKS(4), which makes use of the low-rank properties of local k-space neighborhoods and requires an ACS region to learn the required nullspace filters from submatrices that are fully-sampled, however, calibrationless methods have also been proposed(5)(6)(7)(8).

In addition, the adoption of deep learning techniques for MRI reconstructions (9)(10) is making progress. Specifically, GRAPPA has been generalized using the deep learning method RAKI (Robust Artificial Neural Networks for k-space Interpolation)(11) to enhance noise resilience. RAKI represents a convolutional neural network (CNN) with intermediate ('hidden') layers and nonlinear activations, and can be seen as the nonlinear extension of GRAPPA within a neural network architecture. RAKI's network parameters (i.e. CNN filters) are calibrated using scan-specific ACS as training data. A remarkable feature of RAKI is the strong noise resilience in comparison to GRAPPA, leading to enhanced image reconstruction quality and metrics. In this regard, it is worth noting that analogously to the relationship between RAKI and GRAPPA, the LORAKI(12) method has been proposed to extend AC-LORAKS by the introduction of nonlinear k-space relationships.

However, while it is known that the noise resilience in images reconstructed by RAKI or LORAKI is linked to the nonlinear activations in k-space, a deeper understanding of this relationship remains to be elucidated. This is an important aspect towards clinical translation, identifying and mitigating potential implications of the noise suppression feature in RAKI and to achieve further optimizations.

The aim of this work is to leverage the 'k-space deep learning black box' nature by introducing an image space formalism for the RAKI reconstruction process (also known as inference). The effect of multiple, complex-valued convolution layers as well as nonlinear activation functions is analyzed by decomposing the network into human-readable and explainable components. This allows to formulate an analytical expression for quantifying the noise propagation in RAKI and to study the effect of nonlinear k-space activations on the reconstructed images in more detail. It is shown that enhanced noise resilience implies apparent image blurring/contrast loss artifacts, which can be regularized by controlling the degree of nonlinearity in the model. Furthermore, a potential center "autocorrelation" artifact induced by k-space activations is identified which manifests as bright spot in the image center.

In contrast to previous works on noise propagation analysis in deep neural networks(17)(18), our work investigates the special task of scan-specific k-space interpolation, which requires dedicated considerations. Our approach is inspired by the geometry factor (g-factor)(15)(16) in classical linear PI, where analytical g-factor expressions exist for both sensitivity encoding (SENSE) and GRAPPA. A more generalized g-factor adaptation for k-space interpolation using CNNs, however, has not been presented yet and is introduced in this work.

## 2. Theory

After reviewing the RAKI signal processing in k-space(11)(19), the translation of convolution weights and activation functions from k-space to the image domain is described ("image space formalism") and an analytic expression for noise propagation in RAKI ("RAKI g-factor") is introduced.

### 2.1. Review of RAKI

The original RAKI network was formulated as a real-valued variant to deal with complex-valued data. In this work, we employed a complex-valued convolutional network(20)(21) (22).

The notation employed in the following is described in Supporting Material S0. Let $\mathbf{S}^{(0)} \in \mathbb{C}^{n_x \times n_y \times n_c}$ denote the 3D tensor representing the zero-filled, undersampled k-space with $n_{x,y}$ denoting the image dimensions, and $n_c$ the physical coil number. $\mathbf{S}^{(0)}$ is processed sequentially through hidden layers for abstract feature extraction (Figure 1). The signal of the $k$th hidden layer $\mathbf{S}^{(k)} \in \mathbb{C}^{n_x \times n_y \times n_c^{(k)}}$ is obtained via a subsequent convolution and nonlinear activation operation

$$\mathbf{S}^{(k)} = \mathbb{C}\mathrm{LReLU}\big(\mathbf{S}^{(k-1)} \circledast \mathbf{W}^{(k)}\big), \qquad k = 1, \dots, n_{hid} \quad \mathbf{(1)}$$

where $\circledast$ denotes a complex-valued convolution, $n_{hid}$ is the total number of hidden layers and $\mathbf{W}^{(k)} \in \mathbb{C}^{b_x^{(k)} \times b_y^{(k)} \times n_c^{(k-1)} \times n_c^{(k)}}$ is a 4D tensor representing the convolution kernel of the $k$th hidden layer with $[b_x^{(k)}, b_y^{(k)}]$ denoting the kernel size (i.e. convolution filter size) assigned to the $k$th hidden layer, and $[n_c^{(k-1)}, n_c^{(k)}]$ denoting the channel numbers assigned to the $(k-1)$th and $k$th hidden layers, respectively. The convolution kernel is dilated by the undersampling rate in Phase encoding (PE) direction. RAKI typically employs two hidden layers ($n_{hid} = 2$). As activation function, Akçakaya et al. used the **Re**ctifier **L**inear **U**nit (ReLU) in the original model(11). In this work, we employ the more general leaky variant of ReLU, where an additional hyperparameter $a \in [0,1]$ defines the slope in the negative part ("negative slope parameter") (Figure 1):

$$\mathrm{LReLU}(q) = \begin{cases} q & \text{for } q \geq 0 \\ a * q & \text{otherwise} \end{cases} \quad \mathbf{(2)}$$

Note that ReLU is a special case of LReLU where $a = 0$, and all following considerations and derivations are also valid for ReLU activation when setting $a = 0$. The negative slope parameter $a$ controls the degree of nonlinearity in the model, which can be used as regularization that controls the trade-off between residual artifacts and noise resilience in the case of limited training data as shown in this work (see sections 3.7 and 4.5.)

The proposed network is complex-valued, we employ the **c**omplex LReLU ($\mathbb{C}$LReLU)(23) where the LReLU is applied separately to the real- and imaginary part of the signal to be activated $\mathbf{S}'^{(k)}$:

$$\mathbf{S}^{(k)} = \mathbb{C}\mathrm{LReLU}\left(\mathbf{S}'^{(k)}\right) = \mathrm{LReLU}\left(\mathrm{Re}\left\{\mathbf{S}'^{(k)}\right\}\right) + \boldsymbol{i}\,\mathrm{LReLU}\left(\mathrm{Im}\left\{\mathbf{S}'^{(k)}\right\}\right) \quad (\mathbf{3})$$

where $\boldsymbol{i}$ is the imaginary number**.**

The final layer performs the actual interpolation step on the nonlinear transformation via hidden layers of the undersampled, multicoil k-space $\mathbf{S}^{(0)}$. Thus, it is activated with the identity operator $\mathrm{id}(\mathbf{X}) = \mathbf{X}$

$$\mathbf{S}^{(\mathrm{int})} = \mathrm{id}\left(\mathbf{S}^{(n_{hid})} \circledast \mathbf{W}^{(\mathrm{int})}\right) = \mathbf{S}^{(n_{hid})} \circledast \mathbf{W}^{(\mathrm{int})} \quad (\mathbf{4})$$

where tensor $\mathbf{S}^{(n_{hid})} \in \mathbb{C}^{n_x \times n_y \times n_c^{(n_{hid})}}$ represents the signal in the final hidden layer (i.e., the nonlinear transformation of $\mathbf{S}^{(0)}$ via hidden layers), tensor $\mathbf{W}^{(\mathrm{int})} \in \mathbb{C}^{b_x^{(\mathrm{int})} \times b_y^{(\mathrm{int})} \times n_c^{(n_{hid})} \times R}$ is the convolution kernel of the final layer of the neural network, and tensor $\mathbf{S}^{(\mathrm{int})} \in \mathbb{C}^{n_x \times n_y \times R}$ denotes the interpolated k-space signals. $R$ denotes the undersampling rate, meaning that outside of the ACS region, only every $R$th phase encoding line is sampled. In this work, all coils are interpolated simultaneously for time benefit, in contrast to the original reported RAKI model where the coils were interpolated independently.

During training, the convolution kernels $\mathbf{W}^{(1,\ \cdots\ ,n_{hid})}$, which parameterize the nonlinear transformation $\mathbf{S}^{(n_{hid})}$ of the original multicoil data $\mathbf{S}^{(\mathbf{0})}$ via hidden layers, and $\mathbf{W}^{(\mathrm{int})}$, which parameterizes the interpolation step on $\mathbf{S}^{(n_{hid})}$, are calibrated jointly in a scan-specific, data-driven manner using ACS by minimizing the $L_2$ error between interpolated and ground truth signals. RAKI is described as a scan-specific reconstruction as the CNN filters are trained for each scan separately.

### 2.2. Image space formalism of RAKI

An integral part of the concept presented in this work is to transfer the individual, complex-valued building blocks of RAKI (Eq. (1)) into an image space formalism. According to the convolution theorem, the action of the convolution to the k-space signal $\mathbf{S}^{(k-1)}$ in Eq. (1) is an elementwise multiplication in the image space. This has previously been demonstrated for GRAPPA reconstructions(16)(24). However, transferring the action of the nonlinear activation requires dedicated considerations: Let $\mathbf{S}'^{(k)}$ denote the signal in the $k$th hidden layer prior to activation. As can be seen from Eqs. (2) and (3), the application of $\mathbb{C}$LReLU leaves real and imaginary parts of signals in $\mathbf{S}'^{(k)}$ greater than zero unchanged, otherwise, scales with the negative slope parameter $a$. Thus, the application of $\mathbb{C}$LReLU in k-space can be viewed equivalently as an elementwise multiplication of $\mathbf{S}'^{(k)}$ with an introduced *activation mask* $\mathbf{A}^{(k)} \in \mathbb{C}^{n_x \times n_y \times n_c^{(k)}}$ (Figure 2A):

$$\mathbf{S}^{(k)} = \mathbb{C}\text{LReLU}\left(\mathbf{S}'^{(k)}\right) = \mathbf{S}'^{(k)} \odot \mathbf{A}^{(k)} \quad \mathbf{(5)}$$

where $\odot$ denotes a complex-valued, elementwise multiplication. The computation of $\mathbf{A}^{(k)}$ can be found in Appendix A1.

By formulating the action of $\mathbb{C}$LReLU to signal $\mathbf{S}'^{(k)}$ as described in Eq. (5), the elementwise multiplication of $\mathbf{S}'^{(k)}$ with $\mathbf{A}^{(k)}$ in k-space can be translated into a corresponding complex-valued convolution operation in image space. Thus, the elementary building block of RAKI in k-space translates into the image domain according to:

$$\mathbb{C}\text{LReLU}\left(\mathbf{S}^{(k-1)} \circledast \mathbf{W}^{(k)}\right) = \left(\mathbf{S}^{(k-1)} \circledast \mathbf{W}^{(k)}\right) \odot \mathbf{A}^{(k)} \underset{\textbf{(I)FFT}}{\Longleftrightarrow} \left(\hat{\mathbf{S}}^{(k-1)} \odot \widehat{\mathbf{W}}^{(k)}\right) \circledast \widehat{\mathbf{A}}^{(k)} \quad \mathbf{(6)}$$

where $\hat{\mathbf{S}}^{(k-1)}$, $\widehat{\mathbf{W}}^{(k)}$ and $\widehat{\mathbf{A}}^{(k)}$ are the inverse Fourier transformations of $\mathbf{S}^{(k-1)}$, $\mathbf{W}^{(k)}$ and $\mathbf{A}^{(k)}$**,** respectively.

Eq. (6) allows the formulation of RAKI reconstruction (i.e. inference) entirely in image space:

$$\hat{\mathbf{S}}^{(k)} = (\hat{\mathbf{S}}^{(k-1)} \odot \widehat{\mathbf{W}}^{(k)}) \circledast \widehat{\mathbf{A}}^{(k)}, \quad \mathbf{(7)}$$

$$\text{and } \hat{\mathbf{S}}^{(\text{int})} = \hat{\mathbf{S}}^{(n_{hid})} \odot \widehat{\mathbf{W}}^{(\text{int})}$$

where tensors $\widehat{\mathbf{W}}^{(1,\ \ldots\ ,n_{hid})}$, $\widehat{\mathbf{A}}^{(1,\ \ldots\ ,n_{hid})}$ and $\widehat{\mathbf{W}}^{(\text{int})} \in \mathbb{C}^{n_x \times n_y \times n_c^{(n_{hid})} \times n_c}$ are the image space reconstruction weights which parameterize the interpolation mapping in image space. Please see Eq. (A2.1) in Appendix A2 for an explicit representation of Eq. (7). When the convolution kernels as well as the activation masks are trained in k-space (Figure 2B) and transferred into the image space, RAKI inference can be performed entirely in the image space following Eq. (7) (Figure 2C).

In the following, the spatial dimensions are merged $n = n_x \times n_y$ for readability, thus, the 3D tensors representing multicoil k-space data ($n_x \times n_y \times n_c$) become 2D matrices ($n \times n_c$). The de-aliased coil images, represented as matrix $\widehat{\boldsymbol{S}}^{(\text{int})} \in \mathbb{C}^{n \times n_c}$**,** are finally combined to the reconstructed image, represented as vector $\hat{\boldsymbol{s}}^{(\text{acc})} \in \mathbb{C}^n$**,** using coil-combination weights, represented as matrix $\boldsymbol{P} \in \mathbb{C}^{n \times n_c}$: $\hat{\boldsymbol{s}}^{(\textbf{acc})} = (\widehat{\boldsymbol{S}}^{(\text{int})} \odot \boldsymbol{P})\boldsymbol{u}$, where $\boldsymbol{u}$ is a $n_c \times 1$ unit column vector. The elements of $\boldsymbol{P}$ are obtained by (16) $P_{lk} = \hat{S}_{lk}^{*(\text{int})} / \hat{s}_l^{(\text{acc})}$, where $\hat{s}_l^{(\text{acc})}$ is voxel $l$ in the de-aliased, coil-combined image, and $\hat{S}_{lk}^{*(\text{int})}$ is the complex-conjugate of voxel $l$ in the de-aliased coil image $k$.

### 2.3. Generalized g-factor formulation

In general, the g-factor, represented as vector $\boldsymbol{g} \in \mathbb{R}^n$**,** is defined as the SNR in the fully-sampled reference image, represented as vector $\boldsymbol{snr}^{(\text{full})} \in \mathbb{R}^n$, divided by the SNR in the de-aliased, coil-combined image $\boldsymbol{snr}^{(\text{acc})} \in \mathbb{R}^n$ (15)(16). For voxel $l$, it reads explicitly

$$g_l = \frac{snr_l^{(\text{full})}}{snr_l^{(\text{acc})}\sqrt{R}} = \frac{\sqrt{var_l^{(\text{acc})}}}{\sqrt{var_l^{(\text{full})}}\sqrt{\text{R}}} \quad \mathbf{(8)}$$

where vectors $\boldsymbol{var}^{(full/acc)} \in \mathbb{R}^N$ represent the variance of the reference and the de-aliased, coil-combined image, respectively. Note that Eq. (8) holds under the assumption of similar mean signal in the fully sampled reference and accelerated image, i.e. the latter is an unbiased estimate of the reference image. For the datasets investigated in this work, this assumption is met across a range of noise levels (see Figures S14 and S15). $var_l^{(acc)}$ can be estimated in first order approximation (14) using the Jacobian of the de-aliased, coil-combined image relative to the aliased coil images, $\mathbf{J}^{(\mathrm{acc})} = \boldsymbol{\partial}\hat{\boldsymbol{s}}^{(\mathrm{acc})} / \boldsymbol{\partial}\hat{\boldsymbol{S}}^{(0)} \in \mathbb{C}^{n \times n \times n_c}$ : $var_l^{(acc)} = \sum_{m=1}^{n} \sum_{t=1}^{n_c} \mathrm{J}_{l;mt}^{(\mathrm{acc})} \Sigma_{\mathrm{tt}}^2 \mathrm{J}_{l;mt}^{*(\mathrm{acc})}$, with $\boldsymbol{\Sigma}$ denoting the coil covariance matrix; and $\mathrm{J}_{l;mt}^{(\mathrm{acc})}$ is the derivative of voxel $l$ in the de-aliased, coil-combined image relative to voxel $m$ in the aliased coil image $t$.

Thus, for the generalized g-factor of voxel $l$ in the de-aliased, coil-combined image, we obtain:

$$g_l = \sqrt{\frac{\sum_{m=1}^{n} \sum_{t=1}^{n_c} \mathrm{J}_{l;mt}^{(\mathrm{acc})} \Sigma_{\mathrm{tt}}^2 \mathrm{J}_{l;mt}^{*(\mathrm{acc})}}{\sum_{t=1}^{n_c} P_{lt} \Sigma_{\mathrm{tt}}^2 P_{lt}^*} \frac{1}{\sqrt{R}}} \quad (\mathbf{9})$$

where $P_{lt}$ is the coil-combination weight of voxel $l$ in coil $t$ (obtained following Ref. (16), see above).

Using the image space formulation of RAKI (Eq. (7)), it is possible to obtain an algebraic expression of the derivative of the output tensor of each hidden layer relative to the input tensor of that layer. Thus, using the chain rule and the coil-combination weight matrix $\boldsymbol{P}$, we can express the Jacobian of the de-aliased, coil-combined image relative to the aliased coil images, $\mathbf{J}^{(\mathrm{acc})}$, explicitly. For voxel $l$ in the de-aliased, coil-combined image relative to voxel $m$ in the aliased coil image $t$, its elements are

$$\mathrm{J}_{l;mt}^{(\mathrm{acc})} = \frac{\partial \hat{s}_l^{(\mathrm{acc})}}{\partial \hat{S}_{mt}^{(0)}} = \sum_{h=1}^{n_c} \mathrm{J}_{lh;mt}^{(\mathrm{int})} \cdot P_{lh} \quad (\mathbf{10})$$

where $\mathrm{J}_{lh;mt}^{(\mathrm{int})}$ denotes the derivative of voxel $l$ in the interpolated, de-aliased coil image $h$ relative to voxel $m$ in the aliased coil image $t$ (i.e the input to the network in image space). A detailed derivation of Eq. (10) can be found in Appendix A2. Please also note that in the limit of one linear activated convolution layer, as employed in GRAPPA, Eq. (10) results in the well-known GRAPPA g-factor (see Supporting Material S1 for elaboration).

## 3. Methods

### 3.1. Reconstruction network

The RAKI network architecture used in this work is complex-valued(20)(21)(22) and included 2 hidden layers ($n_{hid} = 2$) with channel numbers $n_c^{(1)} = 128$ and $n_c^{(2)} = 64$. In this work, two datasets with $n_c$ = 20 and 16 were considered, see below for details. The final layer is assigned $n_c^{(\mathrm{int})} = R \cdot n_c$ channels, as all coils are interpolated simultaneously in one network. The assigned kernel sizes are $[b_x^{(1)}, b_y^{(1)}]$= [5,2], $[b_x^{(2)}, b_y^{(2)}]$= [1,1] and $[b_x^{(\mathrm{int})}, b_y^{(\mathrm{int})}]$= [5,1]. No bias terms were included, and kernel dilation was used in Phase encoding direction with the undersampling rate $R$ as dilation rate in each layer. Note that the number of trainable parameters depends on the coil-number and undersampling rate. For the investigated datasets (see next section), the number of trainable parameters amounts to 98,304 and 108,544 for FLASH at $R = 4$ and $R = 5$, respectively, and 131,584 and 144,384 for TSE at $R = 5$ and $R = 6$, respectively. We use $a = 0.5$ as the negative slope parameter in $\mathbb{C}\mathbf{LReLU}$, which was chosen based on a trade-off between noise resilience and residual artifacts for limited training data (see sections 3.7. and 4.5.). The ADAM(25) was used as optimizer in training (learning rate 5e-4). The full ACS data was used as training data in each training iteration (i.e. batch-size is one).

All reconstructions were performed offline using the Python language (v3.11.5) within the PyTorch deep learning framework(26) (v2.1.1) and were conducted on a high-performance-computing cluster with Intel XeonGold 6134(CPU, 360GB RAM).

### 3.2. Datasets

Two transversal 2D brain datasets were acquired from healthy volunteers at 3T (Siemens Magnetom Skyra, Siemens Healthineers, Erlangen, Germany) using a 20-channel head–neck coil array. All experimental procedures were in accordance with institutional guidelines and the study complied with the Helsinki Declaration on Human Experimentation. A T1w FLASH was acquired using TR/TE=250/2.9ms, flip angle=70°, FOV=230×230mm², matrix-size=320×320, slice thickness=3mm, activated coils ($n_c$) =16. A T2w TSE was acquired using turbofactor=12, TR/TE=6000/100ms, FOV=256x256mm², matrix size=264x256, slice thickness=2mm, activated coils ($n_c$) =20.

All datasets were retrospectively undersampled outside the ACS region with rate $R$. ACS data were not re-inserted into the reconstructed k-spaces in all cases for better evaluation of reconstruction quality and comparability.

### 3.3. Domain quasi-equivalent inference

Experiments showing the equivalent performance of RAKI inference in both k-space and image domain were performed on the FLASH and TSE dataset at $R = 4 - 6$. Evaluations were performed both qualitatively using error maps and quantitatively using the normalized mean squared error (NMSE), structural similarity index measure (SSIM) (27) and peak SNR (PSNR). We masked all reconstructions to outline the brain structures (masks obtained via ESPIRiT(28)).

As the training data amount is a crucial limitation in RAKI, we tested RAKI in case of extended training data amount (40 ACS lines), and in the case of limited training data amount (14, 18 and 22 ACS lines at $R = 4 - 6$, respectively). As iterative RAKI (iRAKI) was presented as an improvement over standard RAKI in case of limited training data(22), we performed all experiments additionally with iRAKI. The network architecture in iRAKI is kept the same as for standard RAKI, except for the kernel size of the first hidden layer which is increased to $[b_x^{(1)}, b_y^{(1)}]$= [7,4], which results in the number of trainable parameters to be 172,032 and 182,272 for FLASH at $R = 4$ and $R = 5$, respectively, and 223,744 and 236,544 for TSE at $R = 5$ and $R = 6$, respectively.

### 3.4. Analytic g-factor evaluation

Analytic g-factor maps (50x50 matrix size) were compared with noise enhancement maps obtained by Monte Carlo simulations (also known as pseudo-multiple replica approach(29)). The reconstruction was repeated 1,000 times in image space using k-space data superimposed with randomly generated Gaussian noise $\mathrm{N}(\mu = 0, \sigma^2 = 1)$. The k-space data was pre-whitened beforehand. The variance maps were estimated voxel-wise across the image stack, and g-factor maps were obtained according to Eq. (9).

Furthermore, the analytical approach was tested against automatic differentiation, also known as auto differentiation, which is a widely used technique in machine learning. It enables efficient computation of Jacobians in complex mathematical expressions(30). The Jacobians of all voxels in the de-aliased, coil-combined image relative to the aliased coil images were calculated subsequently(14), and the corresponding g-factors were obtained according to Eq. (9).

### 3.5. Noise distribution analysis

Via activation, RAKI may introduce nonlinearities into the reconstruction pipeline, which may lead to non-Gaussian noise propagations. Thus, the validity of the RAKI g-factor as defined in Eqs. (10) (9) needs to be evaluated. To determine if the distribution of voxel-magnitudes in the pseudo-replicas conforms to a normal distribution, the Kolmogorov-Smirnov (KS) normality test was applied with a sample size of 10,000. The KS test evaluates whether a given sample adheres to a normal distribution, which is considered as the null hypothesis(31). A p-value is calculated by comparing the test statistic to a distribution of the KS test statistic under the null hypothesis (significance level 0.05). The FLASH dataset at $R = 4$ and $5$, and the TSE dataset at $R = 5$ and $6$ using 40 ACS lines were used for evaluation.

### 3.6. Base SNR dependence

The analytical g-factor maps were also computed for different base line SNR levels. To reduce the base SNR retrospectively, the pre-whitened raw data was superimposed with Gaussian noise with elevated standard deviations ($\sigma = 3$ and $\sigma = 5$). Noise was added both to the retrospectively undersampled data and to the fully sampled reference data. The g-factor maps were compared to those obtained via Monte Carlo simulations (1,000 repetitions) for the FLASH data at $R = 4$ and the TSE data at $R = 5$ using 40 ACS lines.

### 3.7. The role of nonlinear activations in RAKI

The image noise suppression in RAKI is related to the nonlinear activations of hidden layers in k-space. In the RAKI model used in this work, the leaky $\mathbb{C}\mathbf{ReLU}$ is employed, where the negative slope parameter $a$ controls the degree of nonlinearity in the model. While $a = 1$ results in a linear interpolation model, setting $a = 0$ results in maximum degree of nonlinearity. While maximum nonlinearity promises also maximum noise suppression, the negative slope parameter $a$ needs to be treated as a hyperparameter in the case of limited training data. This is because residual artifacts may arise which can be partially regularized by reducing noise resilience as trade off. To demonstrate this type of regularization, the negative slope parameter was varied between $a = 0.0, 0.1, \ldots, 1.0$ with incremets of $0.1$ at limited training data (e.g. 18 ACS lines at $R = 4$). Error maps as indicator for reconstruction errors and g-factor maps as indicator for the noise resilience were inspected.

Furthermore, activation masks for different degrees of nonlinearity are characterized, and potential implications of the noise suppression feature are described (i.e. autocorrelation artifact and image blurring).

# 4. Results

## 4.1. Domain quasi-equivalent inference

Figure 3 depicts GRAPPA and RAKI reconstructions of the FLASH dataset at $R = 4$ and 40 ACS lines (total acceleration 2.9). The noise resilience in RAKI is clearly shown in the corresponding error maps and supported by significantly reduced NMSE and improved SSIM and PSNR metrics compared to GRAPPA. As proposed in this work, the RAKI inference was also performed in the image domain. The corresponding error maps and metrics verify quasi-equivalence to the k-space reconstructions. Minor deviations may be attributed to convolution weights padding, as they similarly also appear in the GRAPPA image domain inferences. A comparison to iRAKI, which shows enhanced reconstruction performance relative to standard RAKI and GRAPPA is shown in Figure S1, where furthermore the imaging scenario at $R = 5$ using 40 ACS lines is included (total acceleration 3.3).

Figure 4 depicts reconstructions obtained via GRAPPA and iRAKI in case of limited training data ($R = 5$ using 18 ACS lines, total acceleration 4.0). iRAKI shows enhanced noise resilience relative to GRAPPA, while also minimizing residual reconstruction errors, which are pronounced in standard RAKI due to limited training data (see Figure S2 for comparisons to standard RAKI reconstructions, and for further reconstructions at $R = 4$ and 14 ACS lines (total acceleration 3.5). In this test case, k-space and image space inference are quasi-identical, too, and residual errors in standard RAKI due to the training data limitation are equally displayed in both k-space and image space inferences, supporting the accuracy of the image space formalism.

In general, the corresponding results of the TSE dataset at $R = 5$ and $R = 6$ for limited and extended training data amount confirm the results obtained from the FLASH dataset. They are shown in the Supporting Material, as for all the following experiments (Figures S7-S13).

## 4.2. Analytic g-factor evaluation

Figure 5A shows low-resolution g-factor maps of GRAPPA and RAKI reconstructions from Figure 3 (FLASH $R = 4$, 40 ACS). The maps obtained through the proposed analytical expression match those obtained through Monte Carlo simulations accurately for GRAPPA and RAKI (Figure 5B). The spatially resolved noise resilience in RAKI compared to GRAPPA is well-characterized, and in agreement with the error maps in Figure 3. The test case at limited training data (Figure 4) is analyzed in Figure 5C and D, clearly demonstrating that the analytical approach for iRAKI aligns closely with the Monte Carlo simulation results.

The RAKI and iRAKI g-factor maps obtained with auto differentiation are identical to those obtained analytically (Figure 5B and 5D). This is expected since both methods compute the same Jacobians. However, there is a significant difference in computation time: The analytical g-factor maps were calculated within 135.7-142.6 seconds in all imaging scenarios for RAKI and iRAKI (0.2 seconds for GRAPPA), while the auto differentiation technique required 5433.6-5506.8 seconds (4.6-5.2 seconds for GRAPPA), resulting in a time cost of $\approx$ 2.2 seconds/voxel ($\approx 0.002$ seconds/voxel for GRAPPA). The Monte Carlo simulations required 701.2-754.8 seconds (5.3-7.6 seconds for GRAPPA) for a total of 1,000 pseudo replicas, resulting in a time cost of $\approx 0.75$ seconds/repetition for RAKI and iRAKI (GRAPPA $\approx$0.007seconds/repetition). Please note that a comprehensive comparison of g-factors for GRAPPA, RAKI and iRAKI is shown in Figures S3 and S4 for extended and limited training data amount, respectively.

### 4.3. Noise distribution analysis

The normal distribution of voxel magnitudes was validated in Monte Carlo simulations for RAKI in comparison to GRAPPA. Figure 6A shows the standard deviation maps for the FLASH dataset at $R = 4$ (40 ACS lines). The KS test was performed by fitting of the voxel magnitudes to a normal function. In Figure 6B, this is shown for four representative voxels located at dedicated points (A-D) in the standard deviation map. At point B, pulsatile blood flow artifacts cause an elevated standard deviation, while at A and C, they are elevated due to the multiple folding of pixels, resulting from the undersampling rate and geometric imaging settings. At D, the standard deviation decreases as the voxels can be de-aliased with high accuracy. However, the histograms of all voxels indicate normal distributions. The p-value maps are shown in the top rows of Figure 6C. In the bottom rows, a binary mask indicates voxels that exhibit p-values greater than 0.05. A normal distribution can be assumed for almost all voxels, which is also shown for $R = 5$ (see Figure S5).

### 4.4. Base SNR dependence

Figure 7 shows RAKI g-factor maps of the FLASH dataset at $R = 4$ (40 ACS lines) which was superimposed with Gaussian noise with increasing standard deviations. The enforced SNR loss is visible in the image reconstructions and error maps (Figure 7A), as well as in the quantitative SNR maps obtained from Monte Carlo simulations (Figure 7D). For those cases, the analytical g-factor maps are still in good agreement with those maps obtained via the Monte Carlo simulations (Figure 7B and C). The results also confirm the so-called GRAPPA paradox and show that higher noise in the ACS has a similar effect like Tikhonov regularization thereby yielding lower g-factors(32).

### 4.5. The role of nonlinear activations

Figure 8A demonstrates how the g-factor estimation can be used to find optimal hyperparameter that controls the noise resilience. The NMSE and the mean g-factor are plotted as functions of the negative slope parameter $a$ in RAKI at $R = 4$ and using only 18 ACS lines as training data. We note that the mean g-factor approaches its minimum at $a = 0.0$ and its maximum at $a = 1.0$, with a linear increase in between. This plot shows that maximum noise resilience is obtained at the maximum degree of nonlinearity. Corresponding g-factor maps and images with error maps are shown in Figure 8C for $a = 0.0$, $0.5$ and $1.0$. At maximum noise resilience ($a = 0.0$), residual artifacts manifesting as blurring and contrast loss become pronounced, leading to a noticeable degradation in reconstruction quality, as indicated by the high NMSE and error maps. When gradually increasing $a$, thereby also decreasing noise resilience, these artifacts are gradually suppressed and are strongly mitigated at $a = 0.5$, where the NMSE reaches a minimum. A further increase in $a$, however, leads to severe noise enhancement, which deteriorates reconstruction quality and NMSE. For $a = 1.0$, which represents a linear model, RAKI converges towards GRAPPA (Figure 8C). This shows that the negative slope parameter $a$ can serve as a Tikhonov-like regularization parameter to control the degree of residual artifacts (i.e., apparent blurring and contrast loss) in trade off for slightly increased noise. A similar behavior is also observed for the iRAKI (Figure 8B and 8C). It is worth noting that the iRAKI strongly suppresses the residual artifacts in standard RAKI also for $a = 0.0$ (maximum noise resilience), while maintaining the excellent suppression of noise enhancement. Image reconstructions for each parameter setting in Figure 8A can be found in Video S1, and the corresponding g-factor maps in Video S2.

### 4.6. Characterization of activation masks and autocorrelation artifact

The relationship between noise resilience in the image reconstructions and activations in k-space can be elucidated in a human-readable way through the image space formalism. In the previous section, it was shown that the noise resilience depends on the negative slope parameter in leaky ReLU. By analyzing the activation masks for the respective cases (Figure 9), this pivotal aspect can be investigated in more detail.

For $a$ =1.0 (minimum noise resilience, linear model), the activation mask in k-space is the unity matrix. Thus, in image space, the corresponding activation mask is a Dirac delta peak in the center (see Figure 9 A for exemplary depiction of activation masks of the center channel of the first layer). A convolution with the Dirac delta peak does not change the signal to be activated, thus, there is no effect of noise resilience, and this RAKI model converges towards GRAPPA. However, by increasing the nonlinearity in the model by decreasing $a$ to 0.5, the activation masks in k-space become dependent on the sign of the real and imaginary parts of k-space signals to be activated. The resulting activation masks in image space can be viewed as scan-specific denoising filters, and the convolution with those denoising filters lead to noise resilience in image space. The maximum degree of nonlinearity is reached by setting $a$ to 0.0. In this case, the signal in the activation masks (i.e. the "denoising signal") is also maximum, and the convolution with the activation mask as denoising step is most effective in terms of noise resilience (see also Figure 9 B for plot of the activation mask in RO direction at center PE line). Corresponding figures for all negative slope parameters in leaky ReLU are included in Video S3 and S4 for RAKI and iRAKI, respectively.

It is worth noting significant implications of enhancing noise resilience. When maximum noise resilience is achieved ($a = 0.0$), the images tend to exhibit increased contamination of neighboring voxels. This contamination is essentially a spread of signal between adjacent voxels, leading to image blurring. The root cause of this blurring lies in the noise suppression step, which involves a convolution operation with activation masks. However, as the convolution operation introduces correlation between voxels, and the activation masks are most pronounced at maximum nonlinearity, this operation can be viewed as an image filter. Thus, images with most noise resilience show elevated image blurring.

Furthermore, it is worth noting that the activation masks reveal a specific pattern. The pattern is four-fold repetitive in PE direction (Figure 9 A). This is because the convolution kernels are dilated in PE direction by the undersampling rate. As the zero-filled, undersampled multicoil k-space serves as input to the network, all intermediate k-spaces within the network are also zero-filled like the original, undersampled k-spaces. Thus, the signals to be activated in image space are also four-fold aliased.

Furthermore, the mask manifests spatial resemblance to the signal to be activated, i.e. it shows a spatial autocorrelation pattern. As a result, the convolution of the signal to be activated with the activation mask results in 'autocorrelation signal peaks' in the activated signal with residual equidistant appearance determined by the undersampling rate. Thus, the final image potentially reveals a signal peak induced by the autocorrelations, which consequently is located in the image center.

Figure 10 exemplary shows a T1-weighted fastMRI dataset(34) revealing this center autocorrelation artifact, which is also observed in similar appearance when using the original RAKI code for reconstruction, too. It is worth noting that the inhouse reconstructions show fewer residual artifacts compared to the reconstructions using the original RAKI code, but slightly more noise enhancement. This is because the inhouse implementation uses as negative slope parameter $a = 0.5$, where the original RAKI code uses the ReLU activation function, i.e. $a = 0$. As shown in the previous section, the negative slope parameter acts as a regularization parameter that trades-off residual artifacts against noise resilience. Furthermore, the autocorrelation artifact, as well as the effect of regularization using the negative slope parameter are recognizable in the original RAKI model compared to the inhouse RAKI model, although the individual implementation differences are: real- vs complex valued network, single- vs multi coil interpolation and channel numbers, Tensorflow vs. PyTorch, real-valued ReLU vs. complex-valued, leaky ReLU. This shows that both the autocorrelation artifact and regularization effect are fundamental, inherent attributes of RAKI, and do not depend on the specific implementation (please refer to Figure S6 for further fastMRI reconstructions with different contrast settings).

## 5. Discussion

This work introduces an image space formalism that transforms the complex-valued inference process from k-space to image space in RAKI. This formalism enables a human-readable decomposition of the RAKI network building blocks and provides the basis for an analytical noise propagation quantification. It was shown that the negative slope parameter of leaky ReLU, which controls the degree of nonlinearity in the interpolation model, can serve as Tikhonov-like regularization. Potential implications of enhanced noise resilience were identified, which were found to be increased image blurring and an autocorrelation artifact.

### 5.1. RAKI image space formalism

The noise resilience in RAKI can be attributed to the activations in k-space, which introduce nonlinearity in the interpolation model. The image space formalism provides a more intuitive, human-readable interpretation of the effect of the activation function. The re-formulation of the activation part as multiplication translates the activation into a convolution in the image space. It is this convolution operation in image space that leads to the noise suppression when compared to GRAPPA. As has been demonstrated, RAKI with linear activation function basically represents an extended GRAPPA with multiple hidden layers but does not yield significantly improved reconstruction performance. However, as also shown in this work, the negative slope parameter $a$ can also serve as Tikhonov-like regularization parameter in RAKI at limited training data to mitigate residual artifacts as trade-off for noise resilience. Learning the activation parameter as optimal, scan-specific control of the trade-off (i.e. scan-specific regularization) remains to be elucidated, e.g. by using Kolmogorov Arnold networks(33) for k-space interpolation.

The presented framework can be extended in the sense that it does not cover all possible tensor operations in the forward pass, but considers convolution and activation operations in k-space. Thus, it is applicable to further k-space interpolation techniques which utilize nonlinear activated convolution layers, like LORAKI(12). Other operations can be included in the forward pass, but this operation must be translated from k-space to image space. For analytical g-factor quantification, the additional operations have also to be included in the computations of the Jacobians using the chain rule. For most operations in k-space, it is known how to translate them to image space, e.g. using the convolution theorem (e.g. for convolutions or multiplications), or the linearity of the Fourier transformation (e.g. for bias terms). However, the scope of this work was to investigate how the activation operation in k-space can be translated into image space, as this aspect is not trivial. Thus, this work provides a foundation for image space formalisms of arbitrary network architectures, and generalization of the framework to incorporate further tensor operations remains to be elucidated in future works.

It is worth noting that the adaptation of the presented image space formalism to arbitrary activation functions beyond leaky ReLU is feasible, as the activation in k-space can always be re-written as an elementwise multiplication with an activation mask. The latter can be obtained simply by dividing the activated signal by the signal to be activated (see Supporting Material S2).

### 5.2. Implications of noise resilience: Image blurring and autocorrelation artifact

The convolution with activation masks as noise suppression step has dedicated implications. As the convolution operation introduces correlation between voxels, this operation can be viewed as an image filter and may come along with potential image blurring or voxel contamination, which effectively may deteriorate contrast fidelity. The framework presented here allows to study the potential impact of the activation function on image blurring as trade-off for noise resilience, as shown in this work, in more detail within each layer and within each channel.

Furthermore, the origin of a potential autocorrelation artifact has been identified in this work. This artifact is induced as the activation masks reveal an autocorrelation pattern to the signal to be activated. It is worth emphasizing that this artifact is occasionally apparent in RAKI reconstructions presented in other works (e.g. see RAKI reconstructions in Figures 2 and 3 in Ref. (35)). However, it appears that this artifact is linked to the nonlinear activation of convolution layers in general, as evidenced by its appearance in LORAKI (see Figures 2 and 6 in Ref. (12)). In principle, the image space formalism should allow for the prediction of this center autocorrelation artifacts by quantifying the degree of autocorrelation of the activation masks and the signals to be activated. While raising awareness is important, strategies for mitigation should be considered. Future studies will focus on the investigation of alternative activation functions potentially exhibiting less autocorrelations to the signals to be activated. Further studies may also focus on the quantification of the image blurring or autocorrelation artifact using the image space formalism to define cost functions penalizing those errors, which already has been proven to be beneficial for noise enhancement(36).

### 5.3. Analytical quantification of noise propagation

Noise propagation in deep neural networks has been investigated in previous works. Semenova et al. proposed an analytical framework for noise propagation in real-valued analog neural networks trained in classification tasks and analog function approximation(17). This framework allows for the SNR estimation of the output neurons in feedforward neural networks by estimating the variance and expectation of each neuron in each layer in the forward pass. However, while this framework is adaptable for DNNs with varying topologies, including convolutional topologies, it does not consider the channel dimension, which is essential in RAKI, where the channel dimension is used to incorporate the multi coil data. In addition, this framework does not handle complex-valued data.

Genzel et al. investigated the impact of statistical noise and adversarial perturbations on the image reconstruction in various end-to-end DNNs(18). The loss of reconstruction accuracy caused by noise was assessed quantitatively by noise-to-error curves. These were generated by plotting the relative noise level against the relative reconstruction error on retrospectively undersampled data. While this work shed light on the noise propagation in end-to-end DNNs, scan-specific neural networks, like RAKI, were not considered. Furthermore, a framework for analytical quantification of the noise propagation was not presented, but the noise enhancement in the reconstructed image was quantified using the relative reconstruction error. However, error maps are not impacted by noise enhancement only, but also through other reconstruction errors, for instance, residual artifacts or blurring.

In contrast to these works, our study considers the special task of k-space interpolation and reformulates the action of the activation part as multiplication with activation masks. This allows the translation of the k-space interpolation to the image space inference using the convolution theorem. The spatial noise variance in the reconstructed image is described analytically using explicit algebraic Jacobians relative to the aliased coil images (input data to the neural network), also considering further processing steps beyond the neural network pass, such as coil combination. The noise propagation through the neural network is determined via the chain rule in the backward pass of the RAKI image space formulation. It is worth noting that in principle, the Jacobian could be computed in k-space, i.e., computing the derivative of each signal in the interpolated, multicoil k-spaces relative to each signal in the undersampled, multicoil k-spaces, and use the (inverse) FFT operators to yield the Jacobians in the image space. However, this approach is not as efficient as computing the Jacobians directly in the image space (see Supporting Material S3 for elaboration), and furthermore, does not allow to visualize the individual building blocks of RAKI in a human readable manner.

The RAKI g-factor may be used as an analytical, fast reconstruction measure when no ground truth references are available. Furthermore, NMSE and SSIM are not specific to noise artifacts, but influenced by other reconstruction errors such as residual aliasing or pulsatile blood flow.

In future studies, the analytical noise enhancement quantification may be combined with quantifications of induced, apparent blurring or contrast loss, that potentially comes along with the noise suppression, as it relies on a convolution in image space. This could lead to a new set of quantitative, exact reconstruction metric maps (i.e. assigning degree of blurring and noise enhancement to each voxel), dedicated to scan-specific image reconstructions with CNNs. Thus, the description of the network inference in the image space opens new paths to gain insights into the functionality of k-space interpolation networks. This may also lead to further optimizations regarding cost functions or network architectures. In a broader context, such metric maps are analogous to so called saliency maps known from deep learning. Saliency maps highlight the most influential regions in an input data point for a model's output(37)(38) and serve as interpretability tools, aiding in the comprehension of complex deep learning models across various applications, including healthcare(39)(40). In the context of image reconstruction, saliency maps have also been used for uncertainty quantification (41)(42)(43)(44)(45)(46)(47)(48).

Compared to Monte Carlo and auto differentiation, the analytical approach is the fastest and most accurate. While auto differentiation is equally accurate, it demands significantly more computation time. However, its advantage lies in its relatively simple implementation. Monte Carlo simulations share this attribute but may also require increased computation time depending on replica number and desired accuracy.

It is important to emphasize that the RAKI g-factor, as defined by Eqs. (8) and (9), relies on two critical conditions during the RAKI inference: (1) normally distributed noise propagation, and (2) an unbiased image reconstruction, which permits the substitution of SNR with variance in the g-factor calculation. Regarding point (1), the assumption of normal noise propagation was verified for the datasets analyzed in this work by applying the Kolmogorov–Smirnov test on 10,000 pseudo replicas generated via Monte Carlo simulation. Concerning point (2), a comparison between RAKI g-factors obtained from SNR maps and from variance maps indicates that there is negligible bias for both GRAPPA and RAKI for the datasets under investigation (see Figures S14 and S15). This justifies RAKI g-factor formulation proposed in this work to estimate of the noise propagation in the inference process. It is, however, important to note that in scenarios involving pre-scan calibration, where the contrast in the training data does not match that of the target data, previous studies have shown that RAKI can exhibit contrast loss artifacts(22). Such artifacts may introduce reconstruction bias, potentially compromising the accuracy of the RAKI g-factor estimation. However, it should also be noted that the iRAKI showed contrast fidelity in the case of pre-scan calibration(22).

Please note that the analytical g-factor realization for RAKI requires explicit activation masks to take the activation part of RAKI into account. The activation masks depend on the specific imaging scan. Consequently, the RAKI g-factor can be regarded as an adaptation of the classical GRAPPA g-factor for convolutional neural networks. It can be seen as alternative technique for SNR estimation, however, in contrast to the GRAPPA g-factor, it does not allow conclusions to be drawn about the coil encoding power. A related issue is that once RAKI was trained on a specific imaging scan within a dynamic imaging series, its weights cannot simply be applied to another frame within that series. This is in contrast to conventional parallel imaging with GRAPPA or SENSE, where calibrated kernels or coil sensitivity profiles can be applied to single frames in that time series, which allows for rapid reconstructions. In future studies, we will investigate how trained RAKI weights can appropriately be adopted to varying coil profiles such that a single pre-trained RAKI network can be applied to a time series.

### 5.4. Limitations

Calculating the Jacobian of the de-aliased, coil-combined image relative to the aliased coil images requires computing intermediate Jacobians on hidden layers. If one layer has 64 channels and the next has 128, the intermediate Jacobian is of size [N, 128, N, 64] and consumes 250-300Gb of memory for low-resolution images (N=50×50). Therefore, memory demand is a limitation of this approach. It is worth noting that in the current implementation, tensors are represented using a 128-bit complex data type (with double precision for both real and imaginary components). While a 64-bit complex data type (single precision for real and imaginary components) would reduce memory usage, it was not employed here, as this data type lacks support for all tensor operations in the PyTorch version used in this study. Alternatively, one could compute Jacobians batch-wise, but this reduces time efficiency compared to Monte Carlo and auto differentiation. Alternatively, reducing image resolution to 32x32 may still adequately captures noise characteristics, with a g-factor map typically calculated within seconds using the analytical approach, and strongly reduced memory demand. Another possibility to reduce memory demand are coil-compression techniques.

In the present study, a first-order linear approximation of the nonlinear interpolation is employed to analyze noise propagation. This approach is considered particularly suitable for providing an initial understanding of the impact of noise under the assumption of small perturbations. The first-order approximation simplifies the analysis and allows for tractable mathematical derivations, offering insights into the behavior of k-space interpolation networks. However, higher-order terms could influence noise propagation. Including such terms in the approximation could refine the accuracy of the noise analysis by capturing more subtle interactions between input perturbations and the nonlinear interpolation. Although this analysis is beyond the scope of the current work, it presents a promising direction for future studies. Future investigations could extend the analysis to second-order or higher-order approximations, potentially leading to a more precise understanding of noise behavior in RAKI.

This work was limited to 2D imaging but can be extended to 3D imaging with two-dimensional undersampling as in CAIPIRINHA (49)(50). However, this demands higher memory costs, and dedicated parallel computation techniques are required.

## 6. Conclusion

An image space formalism of convolutional neural networks for k-space interpolation is introduced enabling analytical quantification of noise propagation and human-readable characterization of the inference process. Specifically, nonlinear activation of k-space signals within CNNs are expressed as elementwise multiplication with an activation mask, which translates into a convolution in the image space. This allows for the formulation of a quasi-identical image space formalism for network inference, where the noise suppression step is attributed to convolutions with scan-specific denoising-filters. The Jacobians of the de-aliased, coil-combined image relative to the aliased coil images can be expressed algebraically. Thus, the noise variance can be calculated analytically in first order approximation. This enables fast and accurate description of the noise characteristics, analogous to g-factor maps in traditional parallel imaging methods.

In addition to noise analysis, the image space formalism provides interpretability by enabling direct inspection of individual RAKI network components in a human-readable manner. This allows for the identification and characterization of dedicated implications of the denoising effect. It turns out that the activation masks, which serve as scan-specific denoising filters, may reveal autocorrelation patterns to the signals to be activated, which potentially cause an artifact in the image center apparent in the form of a bright spot. Further denoising effects by the convolution were identified as apparent image blurring and contrast loss artifacts, which can be visualized using the image space formalism and regularized by controlling the nonlinearity degree of the model.

## Appendix

### A1. Derivation of activation masks

To derive the activation masks $\mathbf{A}^{(k)} \in \mathbb{C}^{n_x \times n_y \times n_c^{(k)}}$ introduced in Eq. (5), *binary masks* $\mathbf{M}_{\mathrm{Re,Im}}^{(k)} \in \mathbb{R}^{n_x \times n_y \times n_c^{(k)}}$ are first defined for the real and imaginary parts of $\mathbf{S}'^{(k)}$ ($\mathrm{Re}\{\mathbf{S}'^{(k)}\}$ and $\mathrm{Im}\{\mathbf{S}'^{(\mathrm{k})}\}$, respectively) according to:

$$\mathbf{M}_{\mathrm{Re,Im}}^{(k)} = \begin{cases} 1 & if \quad \mathrm{Re}\{\mathbf{S}'^{(k)}\},\ \mathrm{Im}\{\mathbf{S}'^{(\mathrm{k})}\} \geq \mathbf{0} \\ a & if \quad \mathrm{Re}\{\mathbf{S}'^{(\mathrm{k})}\},\ \mathrm{Im}\{\mathbf{S}'^{(\mathrm{k})}\} < \mathbf{0}. \end{cases} \quad (A1.1)$$

The activation masks $\mathbf{A}^{(k)}$ are then obtained by

$$\mathbf{S}'^{(k)} \odot \mathbf{A}^{(k)} = \left(\mathbf{M}_{\mathrm{Re}}^{(k)} \odot \mathrm{Re}\{\mathbf{S}'^{(k)}\} + \boldsymbol{i}\, \mathbf{M}_{\mathrm{Im}}^{(k)} \odot \mathrm{Im}\{\mathbf{S}'^{(k)}\}\right)$$

$$\Leftrightarrow \mathbf{A}^{(k)} = \left(\mathbf{M}_{\mathrm{Re}}^{(k)} \odot \mathrm{Re}\{\mathbf{S}'^{(k)}\} + \boldsymbol{i}\, \mathbf{M}_{\mathrm{Im}}^{(k)} \odot \mathrm{Im}\{\mathbf{S}'^{(k)}\}\right) \odot \frac{\mathbf{S}^{*\prime(k)}}{|\mathbf{S}'^{(k)}|^2} \quad (A1.2)$$

where $\odot$ denotes an elementwise multiplication which is complex-valued between tensors $\mathbf{S}'^{(k)}$and $\mathbf{A}^{(k)}$, and real-valued between tensors $\mathbf{M}_{\mathrm{Re,Im}}^{(k)}$ and $\mathrm{Re,Im}\{\mathbf{S}'^{(k)}\}$. $\mathbf{S}^{*\prime(k)}$ is the complex-conjugate of $\mathbf{S}'^{(k)}$.

### A2. Derivation for the Jacobian of the de-aliased, coil-combined image relative to the aliased coil images ($\mathbf{J}^{(\mathrm{acc})}$)

Note that the convolution with the activation mask $\mathbf{A}^{(k)}$ can be viewed as matrix multiplication using a Toeplitz matrix representation of $\mathbf{A}^{(k)}$. Eq. (7) can then be expressed explicitly as

$$\hat{S}_{lh}^{(k)} = \sum_{r=1}^{n} \sum_{k=1}^{n_c^{(k-1)}} \hat{S}_{rk}^{(k-1)} \cdot \widehat{\mathrm{W}}_{rhk}^{(k)} \cdot \widehat{\mathrm{A}}_{lrh}^{(k)} , \qquad (A2.1)$$

$$\text{and } \hat{S}_{lh}^{(\mathrm{int})} = \sum_{k=1}^{n_c^{(n_{hid})}} \hat{S}_{lk}^{(n_{hid})} \cdot \widehat{\mathrm{W}}_{lhk}^{(\mathrm{int})}$$

The Jacobian of the output matrix $\widehat{\boldsymbol{S}}^{(k)}$ of layer number $k$ relative to the input matrix $\widehat{\boldsymbol{S}}^{(k-1)}$ to that layer, $\mathbf{J}^{(k)} = \boldsymbol{\partial}\widehat{\boldsymbol{S}}^{(k)} / \boldsymbol{\partial}\widehat{\boldsymbol{S}}^{(k-1)} \in \mathbb{C}^{n \times n_c^{(k)} \times n \times n_c^{(k-1)}}$, follows from Eq. (A2.1), and its elements are:

$$\mathrm{J}_{lh;mt}^{(k)} = \partial\hat{S}_{lh}^{(k)} / \partial\hat{S}_{mt}^{(k-1)} = \widehat{\mathrm{W}}_{mht}^{(k)} \cdot \widehat{\mathrm{A}}_{lmh}^{(k)} \qquad (A2.2)$$

To obtain the derivative of each voxel in the de-aliased coil images $\widehat{\boldsymbol{S}}^{(\mathrm{int})}$ relative to each voxel in the aliased coil images $\widehat{\boldsymbol{S}}^{(0)}$, i.e. the Jacobian $\mathbf{J}^{(\mathrm{int})} = \boldsymbol{\partial}\widehat{\boldsymbol{S}}^{(\mathrm{int})} / \boldsymbol{\partial}\widehat{\boldsymbol{S}}^{(0)} \in \mathbb{C}^{n \times n_c \times n \times n_c}$, the chain rule leads recursively to:

$$\mathrm{J}_{lh;mt}^{(\mathrm{int})} = \partial\hat{S}_{lh}^{(\mathrm{int})} / \partial\hat{S}_{mt}^{(0)} = \sum_{k=1}^{n_c^{(n_{hid})}} \widehat{\mathrm{W}}_{lhk}^{(\mathrm{int})} \cdot \sum_{z,t} \mathrm{J}_{lk;zt}^{(n_{hid})} \cdot \sum_{e,d} \mathrm{J}_{zt;ed}^{(n_{hid}-1)} \cdot (\dots)$$
$$\cdot \sum_{u,i} \mathrm{J}_{ed;ui}^{(2)} \cdot \mathrm{J}_{ui;mt}^{(1)} \qquad (A2.3)$$

Using the coil-combination weight matrix $\boldsymbol{P}$ and the chain rule, we can finally express the Jacobian of the de-aliased, coil-combined image relative to the aliased coil images, $\mathbf{J}^{(\mathrm{acc})}$. For voxel $l$ in the de-aliased, coil-combined image relative to voxel $m$ in the aliased coil image $n$, its elements are

$$\mathrm{J}_{l;mt}^{(\mathrm{acc})} = \partial\hat{s}_{l}^{(\mathrm{acc})} / \partial\hat{S}_{mt}^{(0)} = \sum_{h=1}^{n_c} \mathrm{J}_{lh;mt}^{(\mathrm{int})} \cdot P_{lh} \quad (A2.4)$$

## Acknowledgements

## Funding Information

## Data availability statement

In the spirit of reproducible research, code of the RAKI image space formalism and g-factor computation is available under the following link: https://github.com/pdawood/imageSpaceRaki

# Figures

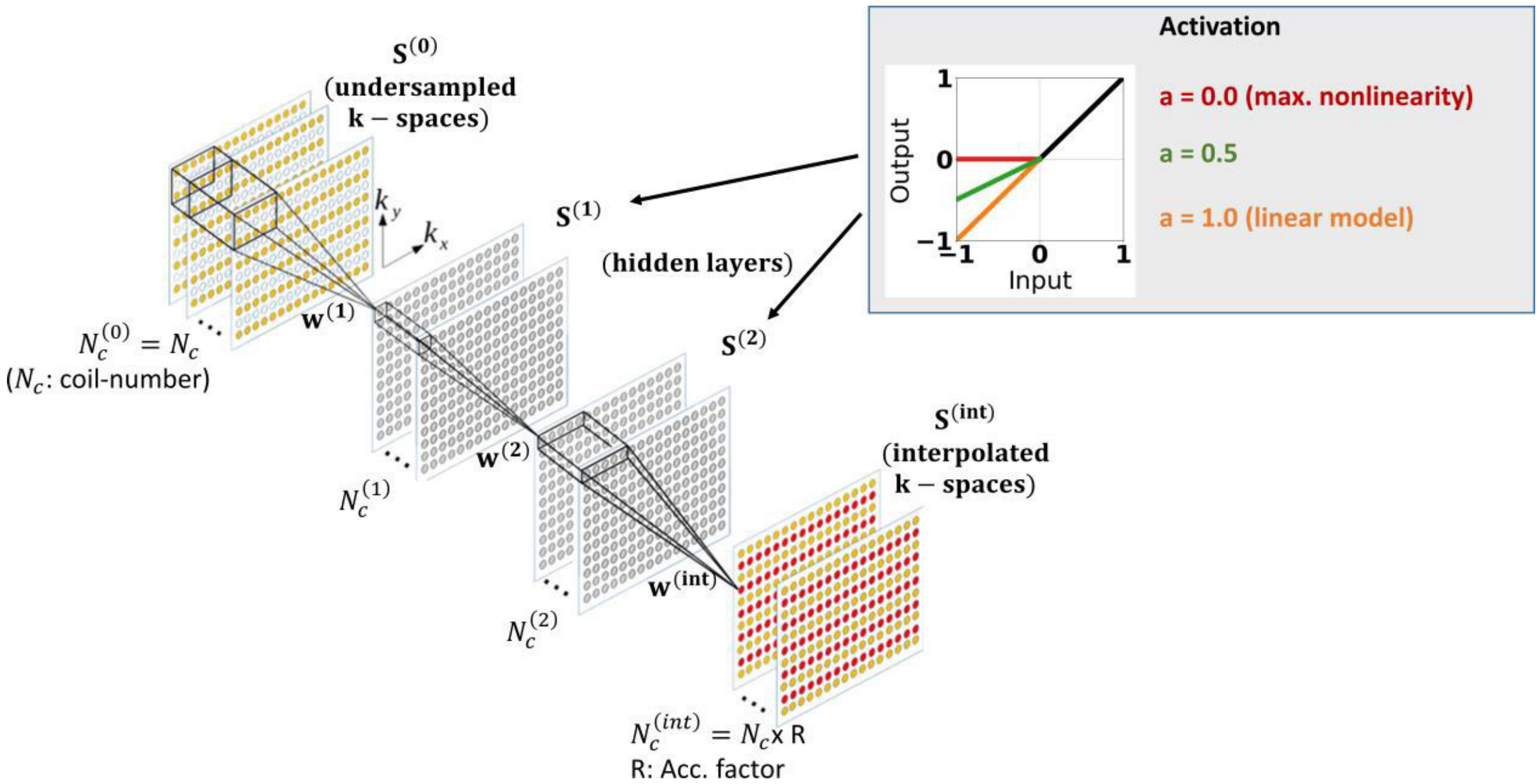

**Figure 1**

The RAKI network architecture employed in this work. The complex-valued convolutional neural network (CNN) takes the zero-filled, undersampled complex-valued, multi-coil k-space as input tensor. Two hidden layers are used for abstract feature extraction with 128 and 64 channels, respectively. The hidden layers process signals by complex-valued convolutions, and nonlinear activation by applying the complex-valued leaky Rectifier Unit (ℂLReLU) elementwise. The degree of nonlinearity in the model is controlled via the negative slope parameter $a$ in ℂLReLU. When $a = 0.0$, maximum nonlinearity is employed, while RAKI becomes a linear model when $a = 1.0$. The CNN outputs the interpolated, complex-valued k-space signals for all coils simultaneously.

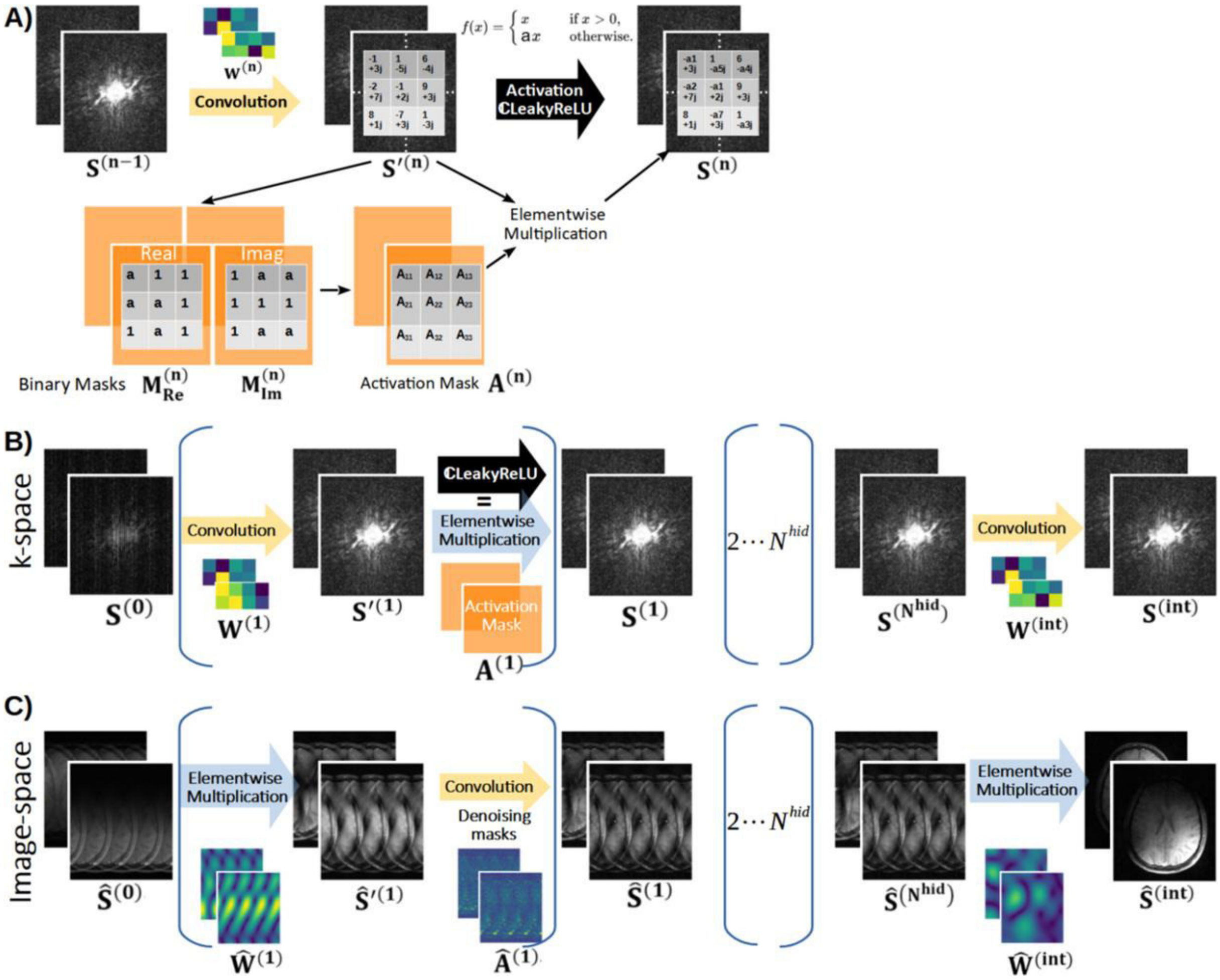

**Figure 2**

**(A)** The proposed formalism to express nonlinear activations in k-space as elementwise multiplication of the signal to be activated with an introduced activation mask $\mathbf{A}$, which is obtained from binary masks $\mathbf{M}$. The latter assign each k-space signal either the value 1 or $a$ (the negative slope parameter of the ℂReLU activation function), depending on the sign of the real and imaginary parts of the signal. **(B)** The conventional RAKI reconstruction in the k-space, in which the training is performed. **(C)** RAKI inference in the image space (proposed method). According to the convolution theorem, convolutions in k-space are transferred into elementwise multiplications in image space, and elementwise multiplications in k-space are transferred into convolutions in image space. All operations are complex-valued.

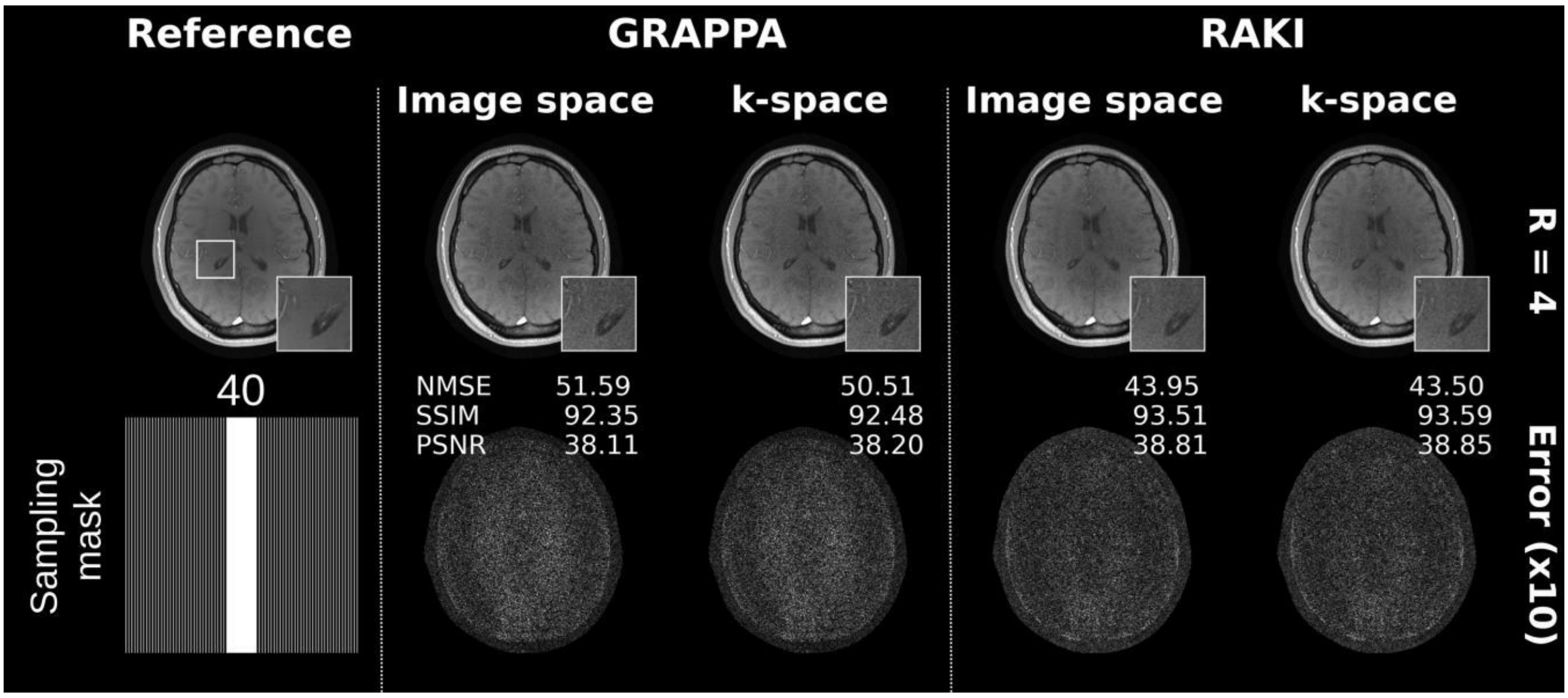

**Figure 3**

**(A)** GRAPPA and RAKI image reconstructions in k-space (conventional method) and in image space (proposed) for FLASH dataset at $R = 4$ using 40 ACS lines as training data, which results in a total acceleration of 2.9. Note that the training takes place in k-space, and only the inference step is performed in image space. The error maps are shown below and scaled for display. Quantitative metrics include the normalized mean squared error (NMSE), structural similarity index measure (SSIM), peak signal to noise ratio (PSNR). Both error maps and quantitative metrics show quasi-identical inference in both domains for all reconstructions. The quasi-identical inference in k-space and image space was previously shown for GRAPPA. In this work, the quasi-identical inference is also shown for RAKI using the proposed image space formalism. Please see Figure S1 for comparison to iRAKI, which shows enhanced reconstruction performance relative to GRAPPA and standard RAKI, and for further reconstructions at $R = 5$ using 40 ACS lines as training data (total acceleration: 3.3).

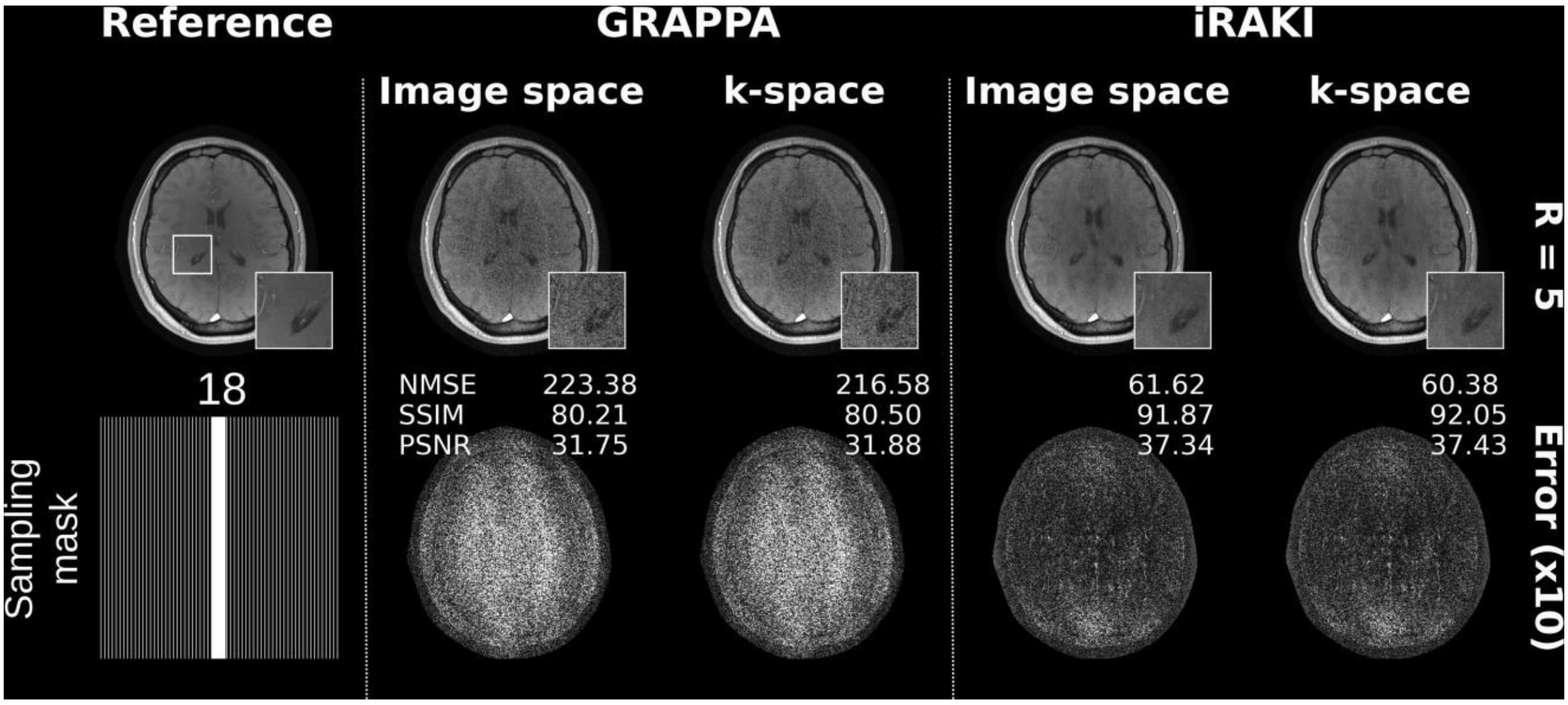

**Figure 4**

**(A)** GRAPPA and iRAKI image reconstructions in k-space (conventional method) and in image space (proposed) for FLASH dataset at $R = 5$ using limited training data amount (only 18 ACS lines, total acceleration 4.0). Error maps are shown below, and scaled for display. Quantitative metrics include the normalized mean squared error (NMSE), structural similarity index measure (SSIM), peak signal to noise ratio (PSNR). Both error maps and quantitative metrics show quasi-identical inference in both domains for all reconstructions, which supports the accuracy of the image space formalism. Please see Figure S2 for the comparison to standard RAKI, which shows severe residual artifacts due to training data limitations, which are equally displayed in the k-space and image space inference, and for further reconstructions at $R = 4$ using only 14 ACS lines as training data (total acceleration 3.5). In all cases, iRAKI outperforms both GRAPPA and standard RAKI by showing suppression of noise enhancement present in GRAPPA, and suppression of residual artifacts present in standard RAKI.

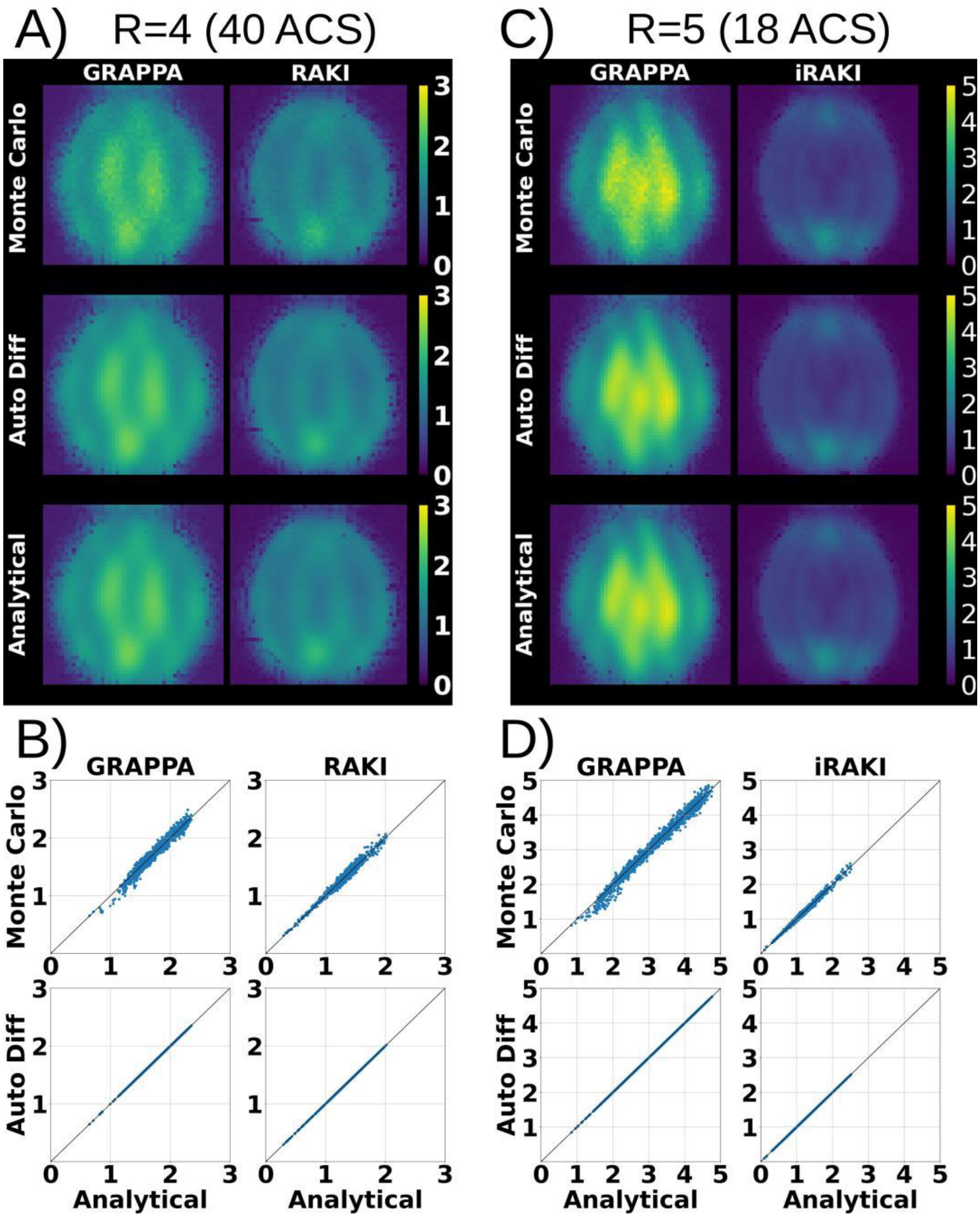

**Figure 5**

**(A)** G-factor maps (50x50 low resolution) computed via Monte Carlo simulations (1,000 repetitions), via auto differentiation and analytically for GRAPPA, RAKI and iRAKI reconstructions (FLASH, $R = 4$, 40 ACS lines). Please note that the analytical g-factor maps for RAKI are calculated the fastest (135.7-142.6 seconds in all imaging scenarios vs. 5433.6-5506.8 seconds by auto differentiation and 701.2-754.8 by Monte Carlo), but demand the most memory usage. For RAKI and iRAKI, it took approx. 262.0 GB ($R = 4$) and 263.0 GB ($R = 5$) (0.5 and 0.7 GB for GRAPPA, respectively), while Monte Carlo took approx. 0.10 GB ($R = 4$) and 0.11 GB ($R = 5$) (0.1 and 0.2 GB for GRAPPA, respectively). G-factor calculations via auto differentiation demanded 6.2 and 6.3 GB for $R = 4$ and $R = 5$, respectively (0.05 and 0.1 GB for GRAPPA). **(B)** G-factors obtained analytically are plotted against those obtained via Monte Carlo (top row) and via auto differentiation (bottom row). The g-factor maps for $R = 5$ and 18 ACS lines are shown in **(C)**, and plots of g-factors obtained analytically against those obtained via Monte Carlo and via auto differentiation are shown in **(D).** Please see Figure S3 for g-factor analysis at $R = 5$ and 40 ACS lines, and Figure S4 for g-factor analysis at $R = 4$ and 18 ACS lines.

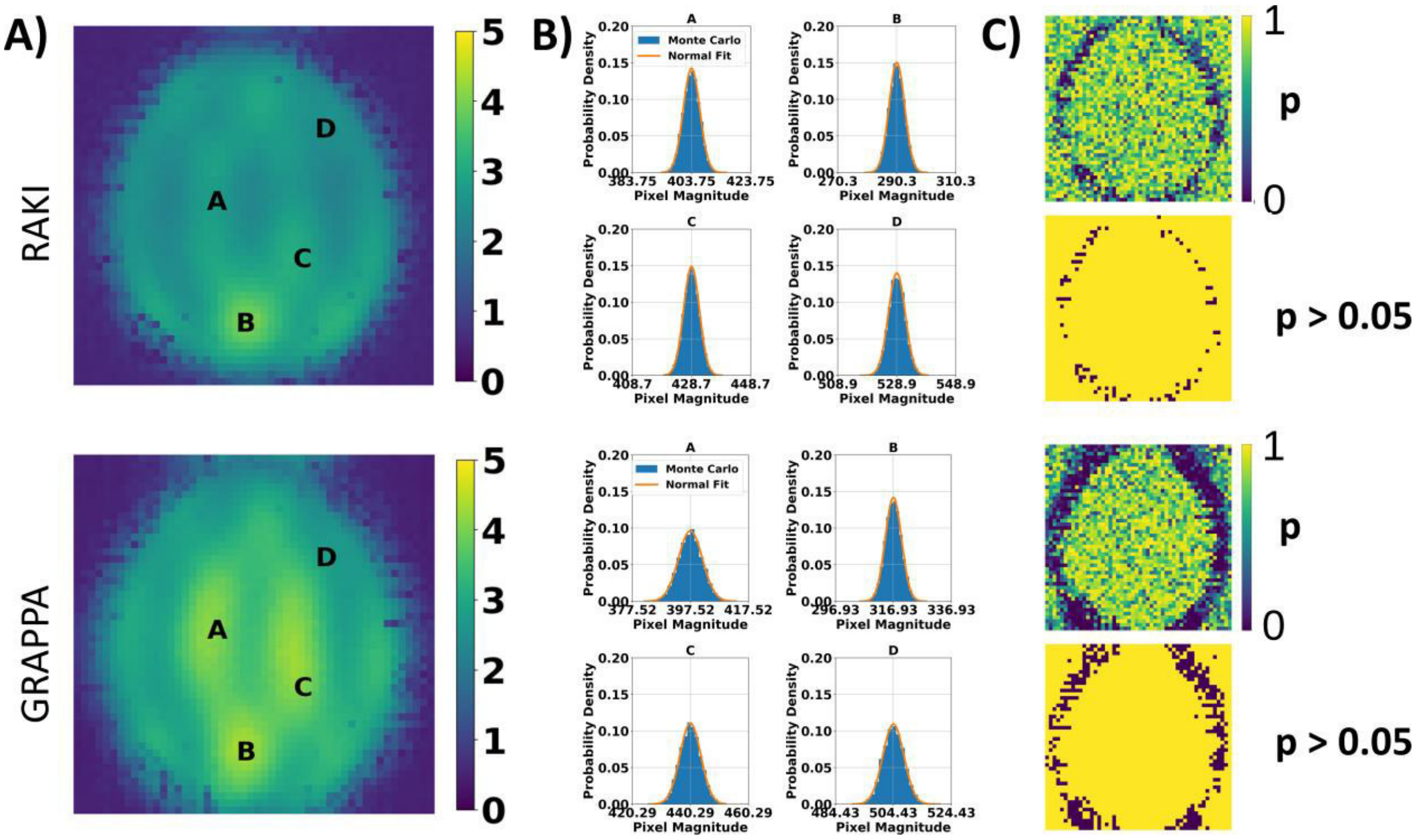

**Figure 6**

**(A)** Standard deviation maps obtained from Monte-Carlo simulations (10,000 repetitions) for RAKI and GRAPPA reconstructions of the FLASH dataset at $R = 4$ (40 ACS lines). **(B)** Voxel magnitude histograms of 10,000 pseudo replicas obtained from voxel locations indexed by A-D in **(A)**, and corresponding fitted normal distributions. **(C)** P-value maps computed in Kolmogorov-Smirnov tests for normality, and binary masks where p>0.05, which is the significance level not to reject the null hypothesis (i.e. voxel magnitude distributions of pseudo replicas are normal). For almost all voxels in the region of interest, a normal distribution can be assumed for both RAKI and GRAPPA, which validates the use of the generalized g-factor computation for RAKI. Please see Figure S5 for $R = 5$.

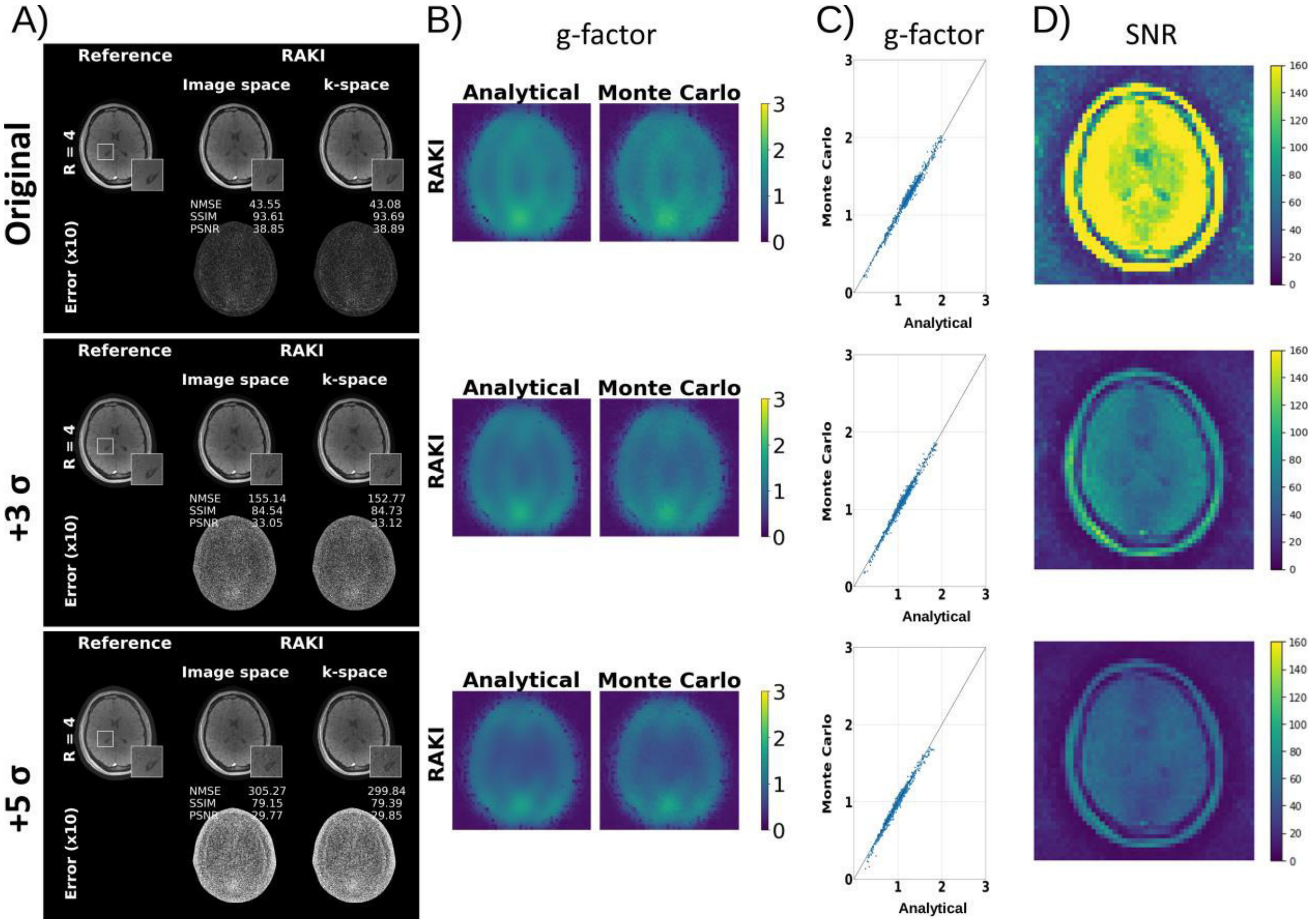

**Figure 7**

**(A)** RAKI image reconstructions and error maps (scaled for display) for the FLASH dataset at $R = 4$ using 40 ACS lines without additive noise superimposition, i.e. original, and with random Gaussian noise superimposition with standard deviation of 3 **σ** and 5 **σ** and zero mean. **(B)** RAKI g-factor maps obtained analytically and via Monte-Carlo simulations (1,000 repetitions). **(C)** G-factors of all voxels computed analytically plotted against g-factors obtained via the Monte Carlo simulations shown in (B). **(D)** SNR maps of images in **(A)** obtained via the Monte-Carlo simulations. Please see Figure S14 for corresponding GRAPPA reconstructions and g-factor maps.

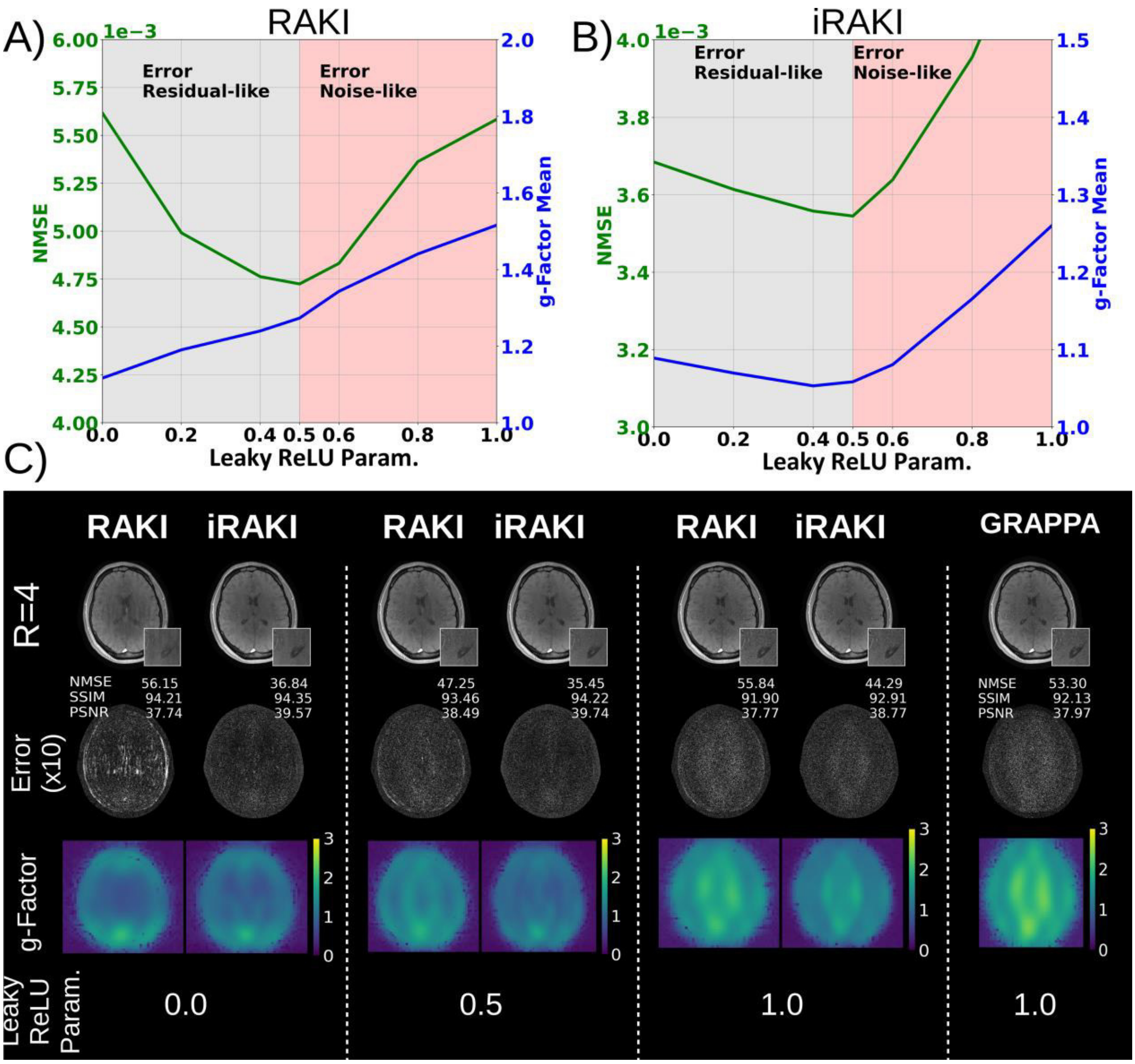

**Figure 8**

**(A)** The normalized mean-squared error (NMSE) and mean g-factor in RAKI (computed analytically) for the FLASH dataset ($R = 4$, 18 ACS lines as training data) as function of the negative slope parameter $a$ in leaky ℂReLU, which controls the degree of nonlinearity in the model. The corresponding image reconstructions, error- and g-factor maps for $a = 0.0, 0.5$ and $1.0$ are depicted in **(C)** in comparison to iRAKI and GRAPPA (linear Model). At maximum nonlinearity ($a = 0.0$), strong noise resilience, but also residual artifacts (blurring and contrast-loss) are present (minimized g-factor, high NMSE). The residual artifacts originate as the denoising effect in image space is a convolution with activation masks, which serve as scan-specific denoising filters. The convolution operation induces voxel correlations, which lead to the blurring and the loss in contrast fidelity. In the linear model ($a = 1.0$), residual artifacts are minimized, however, RAKI suffers from severe noise enhancement (maximum g-factor, high NMSE) as the denoising filters are Dirac-delta peaks, thus, the convolution does not induce voxel correlations and a denoising effect. At medium nonlinearity ($a = 0.5$), a good trade-off between residual artifacts and noise enhancement leads to minimized NMSE, showing that slope $a$ can serve as regularization parameter. **(C)** The NMSE and mean g-factor for iRAKI. It is worth noting that iRAKI strongly suppresses the residual artifacts in standard RAKI also for a=0.0 (maximum noise resilience), while maintaining the excellent suppression of noise enhancement.

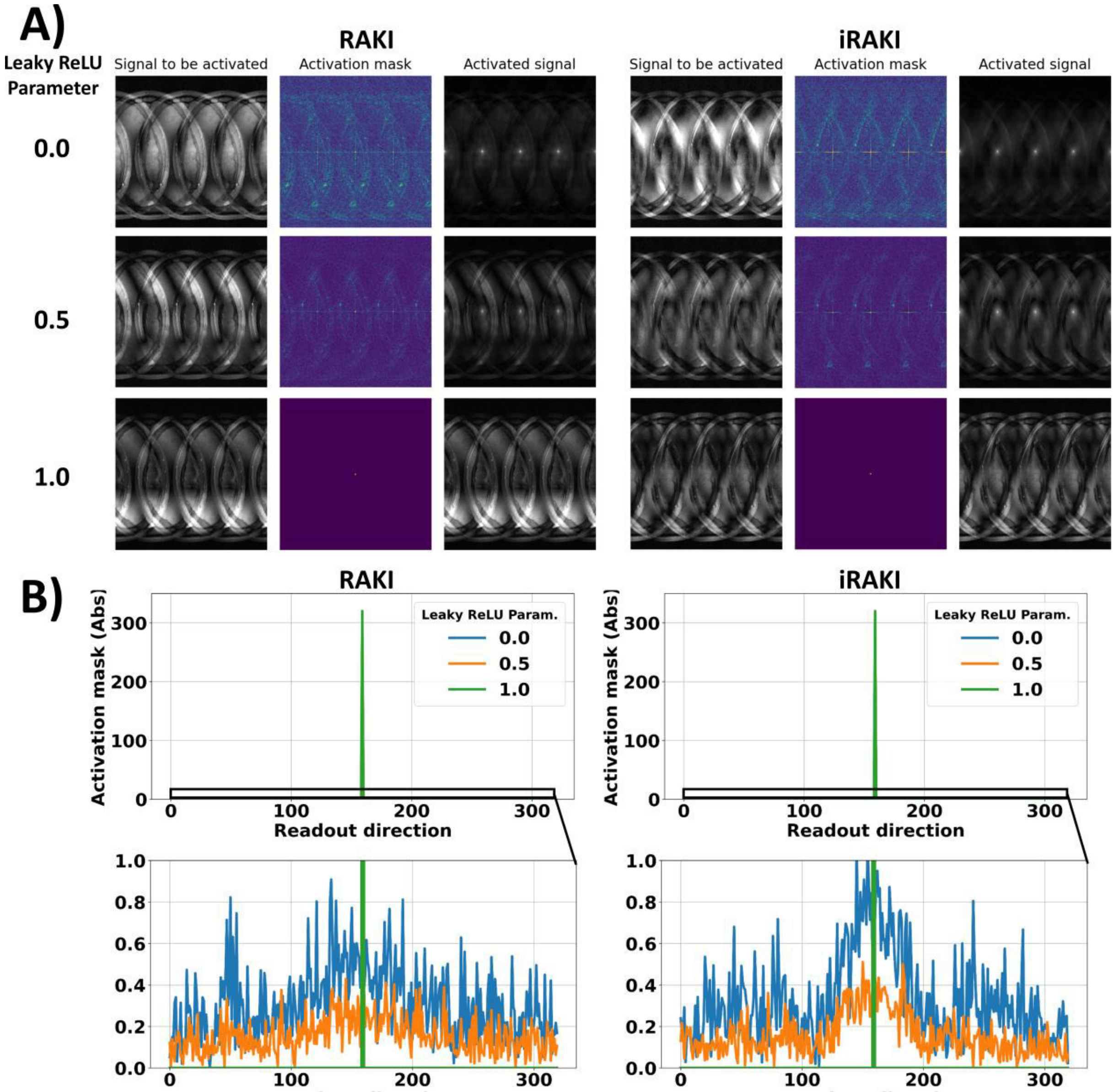

**Figure 9**

**(A)** Activation masks of the center channel in the first hidden layer for RAKI and iRAKI for three degrees of nonlinearity ($a = 0.0$, maximum nonlinear, $a = 0.5$ medium nonlinear, $a = 1.0$ linear model). The convolution kernel is dilated by the undersampling rate $R$ in Phase encoding direction (left to right), hence, the signal to be activated is 4-fold undersampled in that direction. The activated signal is obtained via convolution with activation masks. **(B)** Line plots at center Phase encoding line along Readout direction. Please note the characteristic horizontal stripes along the PE and Readout (RO) direction, which can be attributed to the box-like shape of the activation mask in k-space (alternating between 1 and $a$ for positive and negative sign, respectively). The signal in the activation masks is maximum at maximum nonlinearity ($a = 0.0$=, in this case, the induced correlation between voxels via convolution is most effective for noise resilience. However, the induced correlation leads to blurring-like artifacts due to the broadening of the activation peaks thereby revealing a relationship between image blurring and noise resilience. Decreasing nonlinearity also leads to gradual reduction of magnitudes in the activation masks. Furthermore, the masks manifest spatial resemblance to the signal to be activated. Thus, the convolution of the signal to be activated with the activation mask results in 'autocorrelation signal peaks' in the activated signal. The reconstruction thus may reveal a signal peak, which is always located in the image center (see Figure 10).

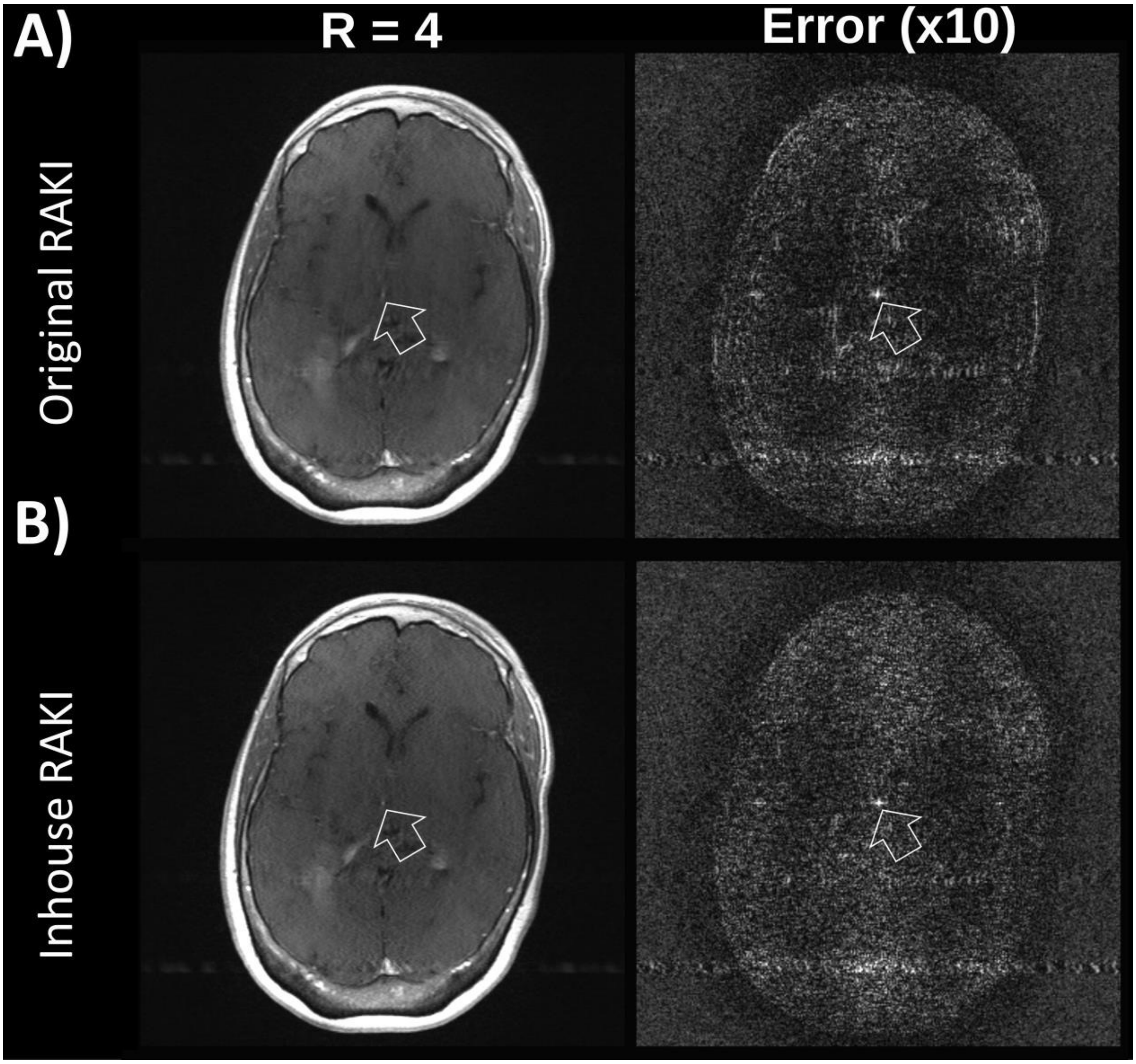

**Figure 10**

A T1-weighted fastMRI in vivo dataset ($R = 4$, 18 ACS lines) reconstructed via original and inhouse RAKI code. Error maps are shown alongside, and scaled for display. In both cases, a pronounced central artifact emerges from the autocorrelation pattern of the activation masks relative to the target signal, as illustrated exemplary in Figure 9. The original RAKI uses ReLU activation (leaky ReLU with negative slope parameter $a = 0.0$), and inhouse RAKI uses leaky ReLU with $a = 0.5$. As demonstrated (Figure 8), parameter $a$ can serve as regularization to trade-off residual artifacts against noise resilience. Thus, original RAKI shows residual artifacts, while inhouse RAKI is slightly noisier with reduced residual artifacts.

# Image space formalism of convolutional neural networks for k-space interpolation

P. Dawood[1,2*], F. Breuer[3], M. Gram[1,4], I. Homolya[5], P.M. Jakob[1], M. Zaiss[2] and M. Blaimer[3]

[1] Experimental Physics 5, University of Würzburg, Würzburg, Germany

[2] Institute of Neuroradiology, University Hospital Erlangen, Erlangen, Germany

[3] Magnetic Resonance and X-ray Imaging Department, Fraunhofer Institute for Integrated Circuits IIS, Division Development Center X-Ray Technology, Würzburg, Germany

[4] Department of Internal Medicine I, University Hospital Würzburg, Würzburg, Germany

[5] Molecular and Cellular Imaging, Comprehensive Heart Failure Center, University Hospital Würzburg, Würzburg, Germany

# Supporting Material

## S0. Notation

**Numbers and Arrays:**

| | |
|---|---|
| $a$ | Scalar (single numerical value, italicized, lower-case letters) |
| $\boldsymbol{a}$ | Vector (1D array, bold, italicized, lower-case letters) |
| $\boldsymbol{A}$ | Matrix (2D array, bold, italicized, upper-case letters) |
| $\mathsf{A}$ | Tensor ($n$D array, $n > 2$, bold sans serif, italicized, upper-case letters) |
| $\mathbf{J}^{(k)}$ | Tensor representing the Jacobian of the output matrix $\boldsymbol{S}^{(k)}$ of layer number *k* relative to the input matrix $\boldsymbol{S}^{(k-1)}$ to that layer ($\mathbf{J}^{(k)} = \boldsymbol{\partial S}^{(k)} / \boldsymbol{\partial S}^{(k-1)}$) |

**Indexing:**

| | |
|---|---|
| $a_i$ | Element *i* of vector $\boldsymbol{a}$ |
| $A_{ij}$ | Element of *i*th row and *j*th column of matrix $\boldsymbol{A}$ |
| $\mathsf{A}_{ijkl}$ | Element along dimensions specified by indices *i,j,k,l* of tensor $\mathsf{A}$ (here, 4-dimensional) |
| $\mathrm{J}^{(k)}_{lh;mt}$ | The derivative of voxel *l* in channel *h* of the output matrix in layer *k* relative to voxel *m* in channel *t* in the input matrix to that layer |

$J^{(\mathrm{int})}_{lh;mt}$ The derivative of voxel *h* of the de-aliased coil image *h* relative to voxel *m* in the aliased coil image *t*

$J^{(\mathrm{acc})}_{l;mt}$ The derivative of voxel l in the de-aliased, coil combined image relative to voxel *m* in the aliased coil image *t*

- The superscript $(k)$ denotes the layer number assigned to scalars, vectors, matrices and tensors
- Scalars, vectors, matrices and tensors with a hat symbol $\hat{\bullet}$ are the corresponding arrays in the image domain, obtained by inverse Fast Fourier transformation $\mathbf{iFFT}\,(\bullet)$:
$$\hat{\mathbf{S}}^{(k)} = \mathbf{iFFT}(\mathbf{S}^{(k)})$$
- Scalars, vectors, matrices and tensors with a star superscript * are complex-conjugate of the corresponding arrays.

## S1. GRAPPA as one-layer special case of RAKI: g-factor calculation

GRAPPA is a special case of RAKI, where no hidden convolution layers are employed. In image space, the interpolated coil images are obtained by elementwise multiplication with the image space GRAPPA kernel $\widehat{\mathbf{W}}^{(\mathrm{int})} \in \mathbb{C}^{n\times n_c\times n_c}$

$$\hat{S}^{(\mathrm{int})}_{lh} = \sum_{t=1}^{n_c} \hat{S}^{(0)}_{lt} \cdot \widehat{\mathrm{W}}^{(\mathrm{int})}_{lht} \quad (\boldsymbol{S1.1})$$

where $\hat{S}^{(0)}_{lt}$ denotes voxel $l$ in the aliased coil image $t$, and $\hat{S}^{(\mathrm{int})}_{lh}$ denotes voxel $l$ in the de-aliased coil image $h$.

The de-aliased, coil-combined image, represented as vector $\hat{\boldsymbol{s}}^{(\mathrm{acc})} \in \mathbb{C}^{n}$, is obtained using the coil-combination weights matrix $\boldsymbol{P} \in \mathbb{C}^{n\times n_c}$:

$$\hat{s}^{(\mathrm{acc})}_{l} = \sum_{h=1}^{n_c} \hat{S}^{(\mathrm{int})}_{lh} \cdot P_{lh} \quad (\boldsymbol{S1.2})$$

where $P_{lh}$ is the coil-combination weight of voxel $l$ in coil $h$.

For the Jacobians of the de-aliased, coil-combined image relative to the aliased coil images, $\mathbf{J}^{(\mathrm{acc})}$, we obtain correspondingly

$$J^{(\mathrm{acc})}_{l;lt} = \frac{\partial \hat{s}^{(\mathrm{acc})}_{l}}{\partial \hat{S}^{(0)}_{lt}} = \sum_{h=1}^{n_c} \widehat{\mathrm{W}}^{(\mathrm{int})}_{lht} \cdot P_{lh} \quad (\boldsymbol{S1.3})$$

and thus, for the g-factor, we obtain

$$g_l = \sqrt{\frac{\sum_{t=1}^{n_c} J^{(\mathrm{acc})}_{l;lt}\, \Sigma^2_{\mathrm{tt}}\, J^{*(\mathrm{acc})}_{l;lt}}{\sum_{t=1}^{n_c} P_{lt} \Sigma^2_{\mathrm{tt}}\, P^*_{lt}}}\, \frac{1}{\sqrt{R}} \quad (\boldsymbol{S1.4})$$

with $\boldsymbol{\Sigma}$ denoting the coil covariance matrix, $\mathrm{J}_{l;lt}^{(\mathrm{acc})}$ is the derivative of voxel $l$ in the de-aliased, coil-combined image relative to voxel $l$ in the aliased coil image $t$. Eq. (S1.4) is the explicit, element-wise expression of the known GRAPPA g-factor.

## S2. Computing activation masks for arbitrary activation functions

Let $\mathbf{S}'^{(n)}$ denote the tensor assigned to the $n$th hidden layer representing the k-space signal to be activated, and let $\sigma$ denote a complex-valued activation function which acts elementwise on $\mathbf{S}'^{(n)}$ such that the activated signal $\mathbf{S}^{(n)}$ is obtained via

$$\mathbf{S}^{(n)} = \sigma\left(\mathbf{S}'^{(n)}\right). \quad (\boldsymbol{S2.1})$$

To rewrite Eq. (***S*2.1**) as an elementwise multiplication ($\odot$) with an activation mask $\mathbf{A}^{(n)}$, i.e.

$$\mathbf{S}^{(n)} = \sigma\left(\mathbf{S}'^{(n)}\right) := \mathbf{S}'^{(n)} \odot \mathbf{A}^{(n)}, \quad (\boldsymbol{S2.2})$$

the activation mask is computed via elementwise division, i.e.

$$\mathbf{A}^{(n)} = \frac{\sigma\left(\mathbf{S}'^{(n)}\right)}{\mathbf{S}'^{(n)}}. \quad (\boldsymbol{S2.3})$$

Eq. (*S*2.2) can then be transformed into image space using the convolution theorem, as an elementwise multiplication translates into a convolution operation.

## S3. Computing $\mathrm{J}^{(\mathrm{int})}$ from Jacobians in k-space

In this work, the variance in the reconstructed image was estimated in first order approximation using the Jacobian of the de-aliased coil images $\widehat{\boldsymbol{S}}^{(\mathrm{int})}$ relative to the aliased coil images $\widehat{\boldsymbol{S}}^{(0)}$, i.e., $\mathrm{J}^{(\mathrm{int})} = \partial\widehat{\boldsymbol{S}}^{(0)}/\partial\widehat{\boldsymbol{S}}^{(\mathrm{int})}$.

To compute $\mathrm{J}^{(\mathrm{int})}$ analytically, the inference process was transformed from k-space to image space in this work (variant 1). Alternatively, $\mathrm{J}^{(\mathrm{int})}$ can be obtained by computing the Jacobian of the interpolated, multicoil k-space data $\boldsymbol{S}^{(\mathrm{int})}$ relative to the undersampled multicoil k-space data $\boldsymbol{S}^{(0)}$, i.e., $\mathbf{J}'^{(\mathrm{int})} = \partial\boldsymbol{S}^{(0)}/\partial\boldsymbol{S}^{(\mathrm{int})}$, and applying the (inverse) Fast Fourier operators (variant 2).

In variant 2, the elements of $\mathrm{J}^{(\mathrm{int})}$ would be obtained explicitly via the chain rule according to

$$\mathrm{J}_{lh;mt}^{(\mathrm{int})} = \frac{\partial\hat{S}_{lh}{}^{(\mathrm{int})}}{\partial\hat{S}_{mt}{}^{(0)}} = \sum_{i=0}^{n}\sum_{v=0}^{n} \frac{\partial\hat{S}_{lh}{}^{(\mathrm{int})}}{\partial S_{ih}{}^{(\mathrm{int})}} \frac{\partial S_{ih}{}^{(\mathrm{int})}}{\partial S_{vt}{}^{(0)}} \frac{\partial S_{vt}{}^{(0)}}{\partial\hat{S}_{mt}{}^{(0)}}$$

$$= \sum_{i=0}^{n}\sum_{v=0}^{n} \mathrm{iFFT}_{li}\ \mathrm{J}_{ih;vt}'^{(\mathrm{int})}\ \mathrm{FFT}_{vm} \quad (\boldsymbol{S3.1})$$

where $n$ is the total voxel number, $\mathrm{iFFT}_{li}$ is the coefficient of the inverse Fourier operator relative to voxel $l$ and k-space signal $i$, $\mathrm{J}_{ih;vt}'^{(\mathrm{int})}$ is the derivative of the $i$th k-space signal in the interpolated coil $h$ relative to the $v$th signal in the undersampled coil $t$, and $\mathrm{FFT}_{vm}$ is the coefficient of the Fourier operator relative to k-space signal $v$ and voxel $m$.

While both variants yield $\mathrm{J}^{(\mathrm{int})}$, there is a significant difference in the relative computational complexity: In variant 1, the relevant operation is a 2D iFFT applied to the k-space data and to the network weights for inference in image space with. For a total voxel number $n$, this is assigned a complexity of $O(n \log n)$. On the other hand, variant 2 requires applying the iFFT and FFT operator on $\mathrm{J}'^{(\mathrm{int})}$ (Eq.S3.1) whose dimension is quadratic in $n$ (contains derivative of each signal in the interpolated k-spaces relative to each signal in the undersampled k-spaces). This is assigned a complexity of $O(n^2 \log n)$. Thus, particularly in 3D imaging where $n$ growths significantly , variant 1 is much more efficient. However, it is worth emphasizing that $\mathrm{J}'^{(\mathrm{int})}$is expected to be sparse, in contrast to $\mathrm{J}^{(\mathrm{int})}$, which in general is dense due to the convolution operation with activation mask in image space in the forward pass. Please note that in case of a sparse $\mathrm{J}'^{(\mathrm{int})}$, the full formation of $\mathrm{J}^{(\mathrm{int})}$ according to Eq. S3.1 still incorporates a complexity of $O(n^2 \log n)$, albeit with a smaller constant factor.

## Supporting Material: Figures

## Figure S1

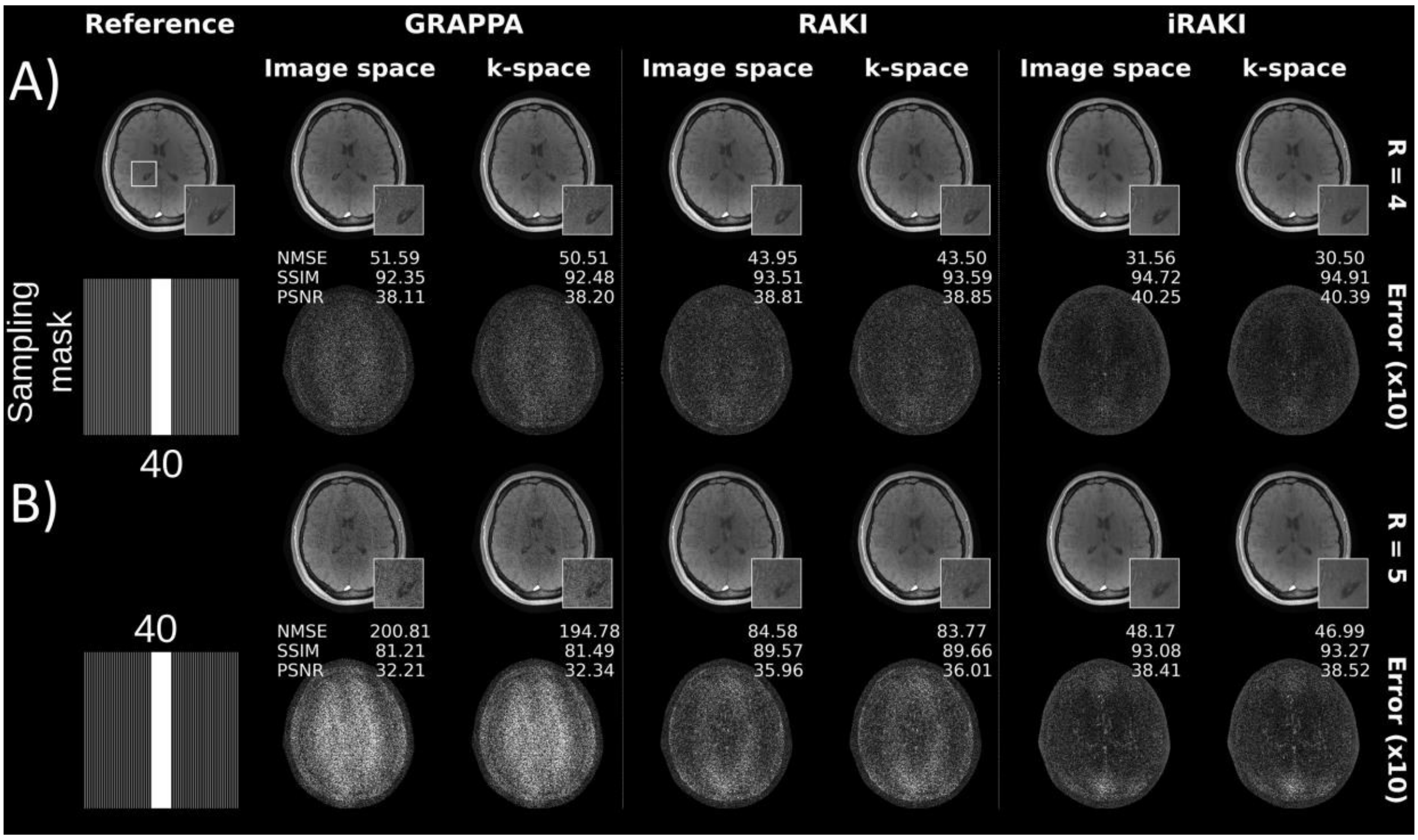

**(A)** GRAPPA, RAKI and iRAKI image reconstructions in k-space (conventional method) and in image space (proposed) for FLASH dataset at $R = 4$ using 40 ACS lines as training data (total acceleration 2.9). Note that the training takes place in k-space, and only the inference step is performed in image space. The error maps are shown below and scaled for display. Quantitative metrics include the normalized mean squared error (NMSE), structural similarity index measure (SSIM), peak signal to noise ratio (PSNR). Both error maps and quantitative metrics show quasi-identical inference in both domains for all reconstructions. The quasi-identical inference in k-space and image space was previously shown for GRAPPA. In this work, the quasi-identical inference is also shown for RAKI and iRAKI using the proposed image space formalism. $R = 5$ (total acceleration 3.3) is shown in **(B)**. Please note, iRAKI yields superior noise resilience and outperforms both GRAPPA and standard RAKI. Imaging scenarios at limited training data and enhanced total accelerations are shown in Figure S2.

## Figure S2

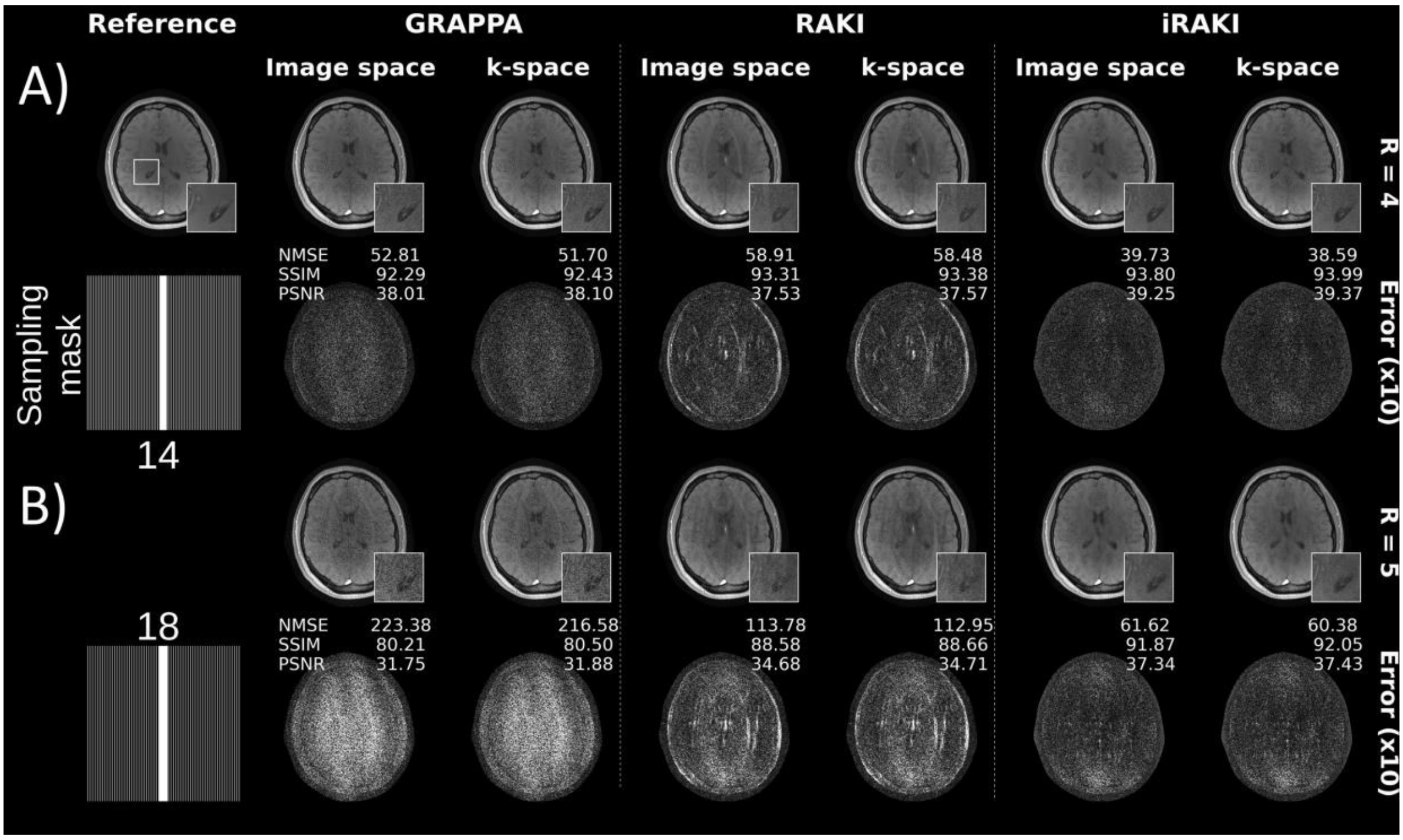

**(A)** GRAPPA, RAKI and iRAKI image reconstructions in k-space (conventional method) and in image space (proposed) for FLASH dataset at $R = 4$ using limited training data amount (only 14 ACS lines, total acceleration 3.5). Residual errors due to the training data limitation are equally displayed in both k-space and image space, supporting the accuracy of the image space formalism. It is worth noting that the iRAKI suppresses residual artifacts in RAKI while providing a similar noise suppression feature, and it shows similar performance as standard RAKI trained with 40 ACS lines (see Figure S1 for the latter). $R = 5$ and using only 18 ACS lines (total acceleration 4.0) is shown in **(B)**.

## Figure S3

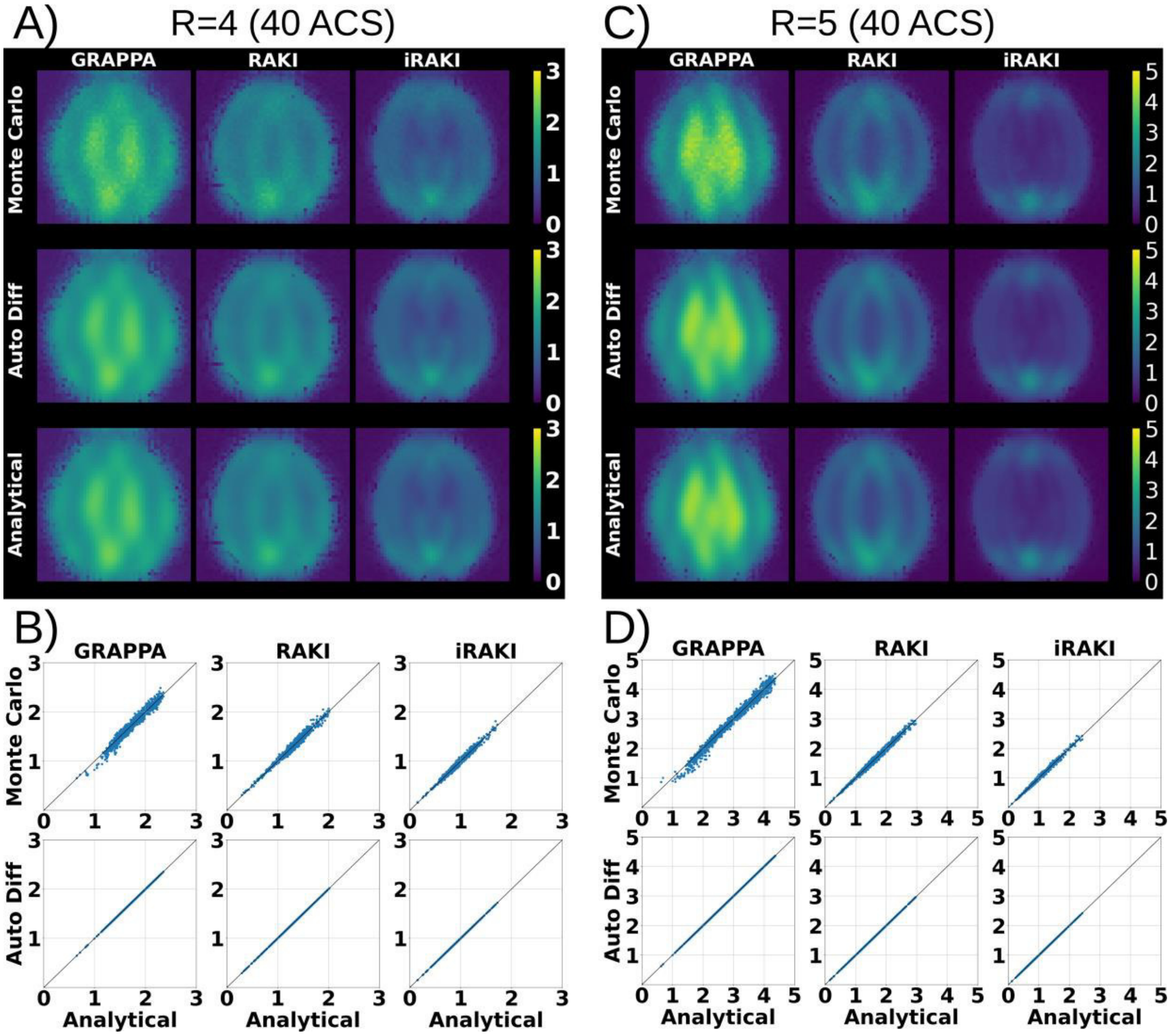

**(A)** G-factor maps (50x50 low resolution) computed via Monte Carlo simulations (1,000 repetitions), via auto differentiation and analytically for GRAPPA, RAKI and iRAKI reconstructions (FLASH, $R = 4$, 40 ACS lines). Please note that the analytical g-factor maps for RAKI are calculated the fastest (135.7-142.6 seconds in all imaging scenarios vs. 5433.6-5506.8 seconds by auto differentiation and 701.2-754.8 by Monte Carlo), but demand the most memory usage. For RAKI and iRAKI, it took approx. 262.0 GB ($R = 4$) and 263.0 GB ($R = 5$) (0.5 and 0.7 GB for GRAPPA, respectively), while Monte Carlo took approx. 0.10 GB ($R = 4$) and 0.11 GB ($R = 5$) (0.1 and 0.2 GB for GRAPPA, respectively). G-factor calculations via auto differentiation demanded 6.2 and 6.3 GB for $R = 4$ and $R = 5$, respectively (0.05 and 0.1 GB for GRAPPA). **(B)** G-factors obtained analytically are plotted against those obtained via Monte Carlo (top row) and via auto differentiation (bottom row). The g-factor maps for $R = 5$ and 40 ACS lines are shown in **(C)**, and plots of g-factors obtained analytically against those obtained via Monte Carlo and via auto differentiation are shown in **(D).**

**Figure S4**

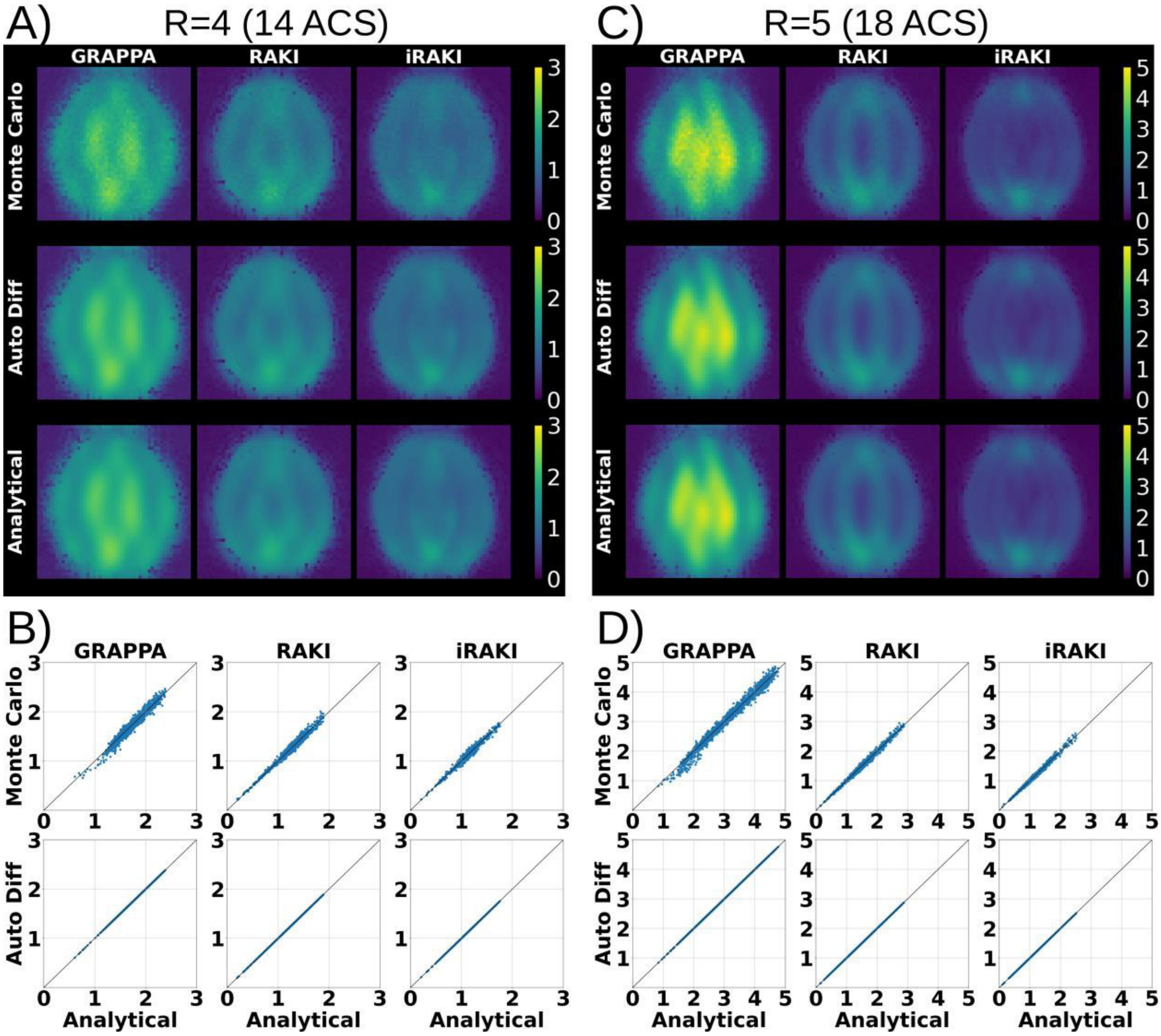

**(A)** G-factor maps (50x50 low resolution) computed via Monte Carlo simulations (1,000 repetitions), via auto differentiation and analytically for GRAPPA, RAKI and iRAKI reconstructions (FLASH, $R = 4$, 14 ACS lines). **(B)** G-factors obtained analytically are plotted against those obtained via Monte Carlo (top row) and via auto differentiation (bottom row). The g-factor maps for $R = 5$ and 18 ACS lines are shown in **(C)**, and plots of g-factors obtained analytically against those obtained via Monte Carlo and via auto differentiation are shown in **(D).**

## Figure S5

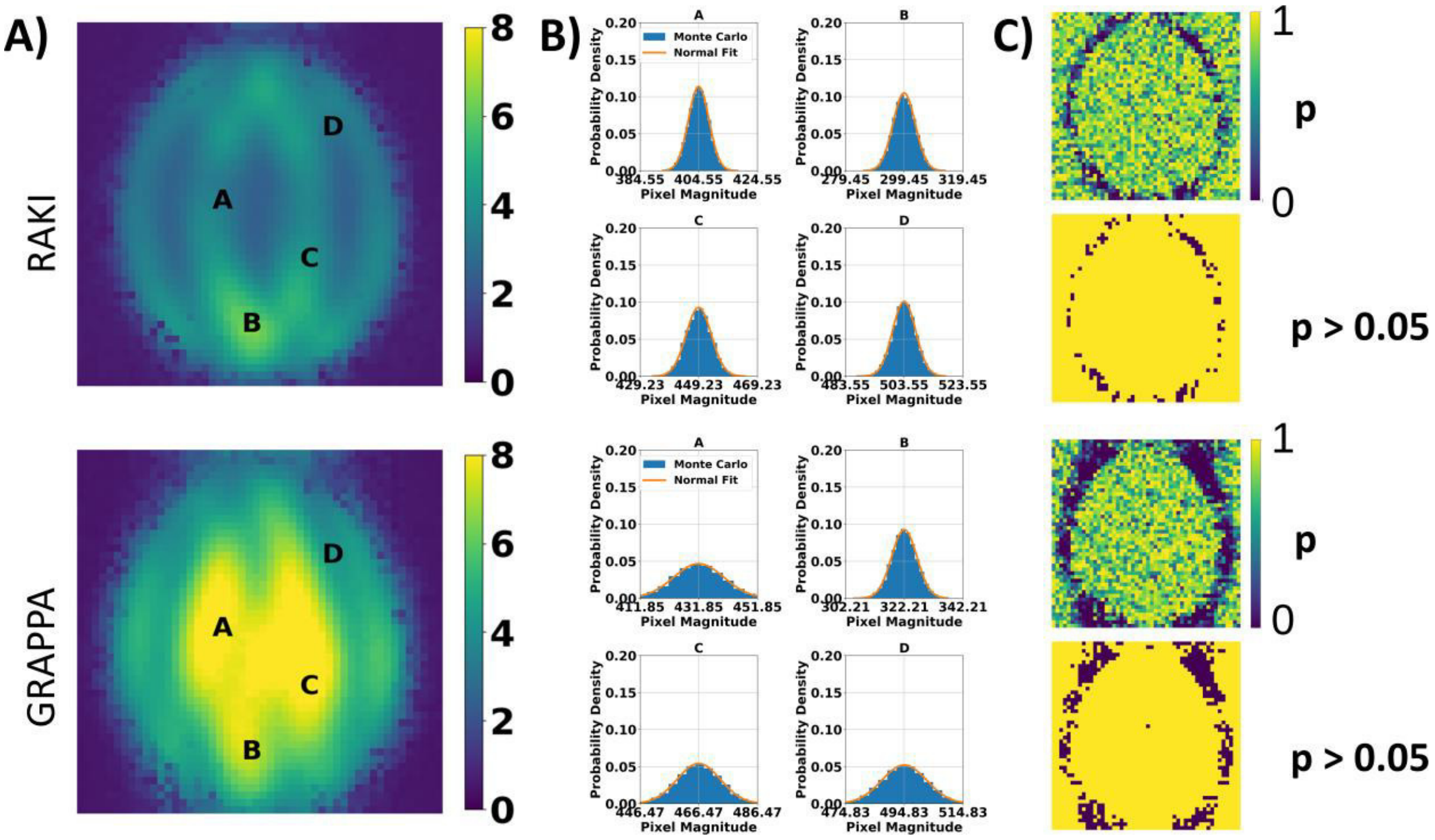

**(A)** Standard deviation maps obtained from Monte-Carlo simulations (10,000 repetitions) for RAKI and GRAPPA reconstructions of the FLASH dataset at $R = 5$ (40 ACS lines). **(B)** Voxel magnitude histograms of 10,000 pseudo replicas obtained from voxel locations indexed by A-D in **(A)**, and corresponding fitted normal distributions. **(C)** P-value maps computed in Kolmogorov-Smirnov tests for normality, and binary masks where p>0.05, which is the significance level not to reject the null hypothesis (i.e. voxel magnitude distributions of pseudo replicas are normal). For almost all voxels in the region of interest, a normal distribution can be assumed for both RAKI and GRAPPA, which validates the use of the generalized g-factor computation for RAKI.

**Figure S6**

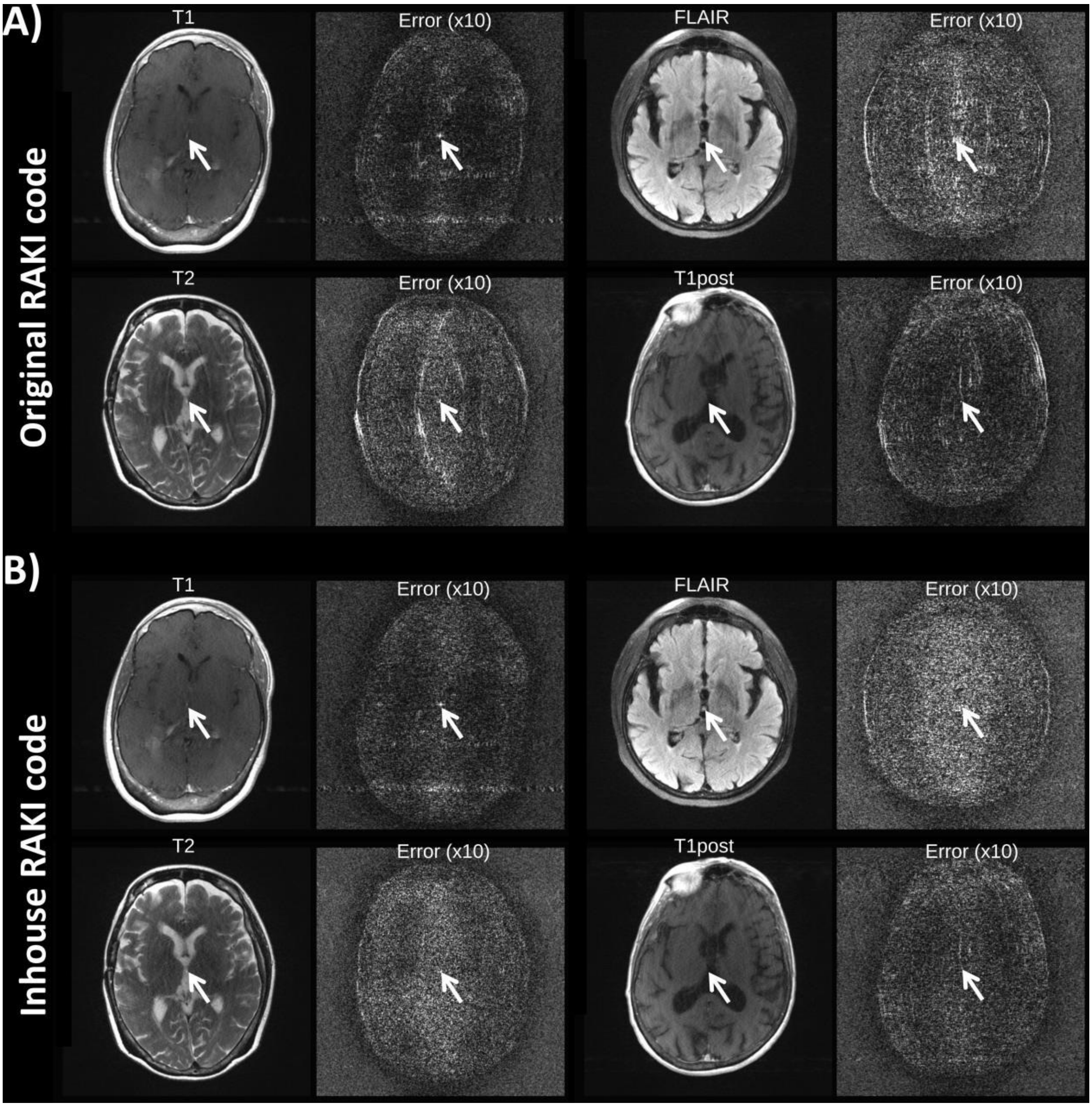

fastMRI in vivo datasets with T1, FLAIR, T2 and T1post weighting ($R = 4$, 18 ACS lines) reconstructed via inhouse and original RAKI code, and error maps shown below (scaled for display). In all cases, a pronounced central artifact emerges from the autocorrelation pattern of the activation masks relative to the target signal, as exemplary illustrated in Figure 9. The original RAKI uses ReLU activation (leaky ReLU with negative slope parameter $a = 0.0$), and inhouse RAKI uses leaky ReLU with $a = 0.5$. As demonstrated (Figure 8), parameter $a$ can serve as regularization to trade-off residual artifacts against noise resilience. Thus, original RAKI shows residual artifacts, while inhouse RAKI is slightly noisier with reduced residual artifacts.

## Figure S7

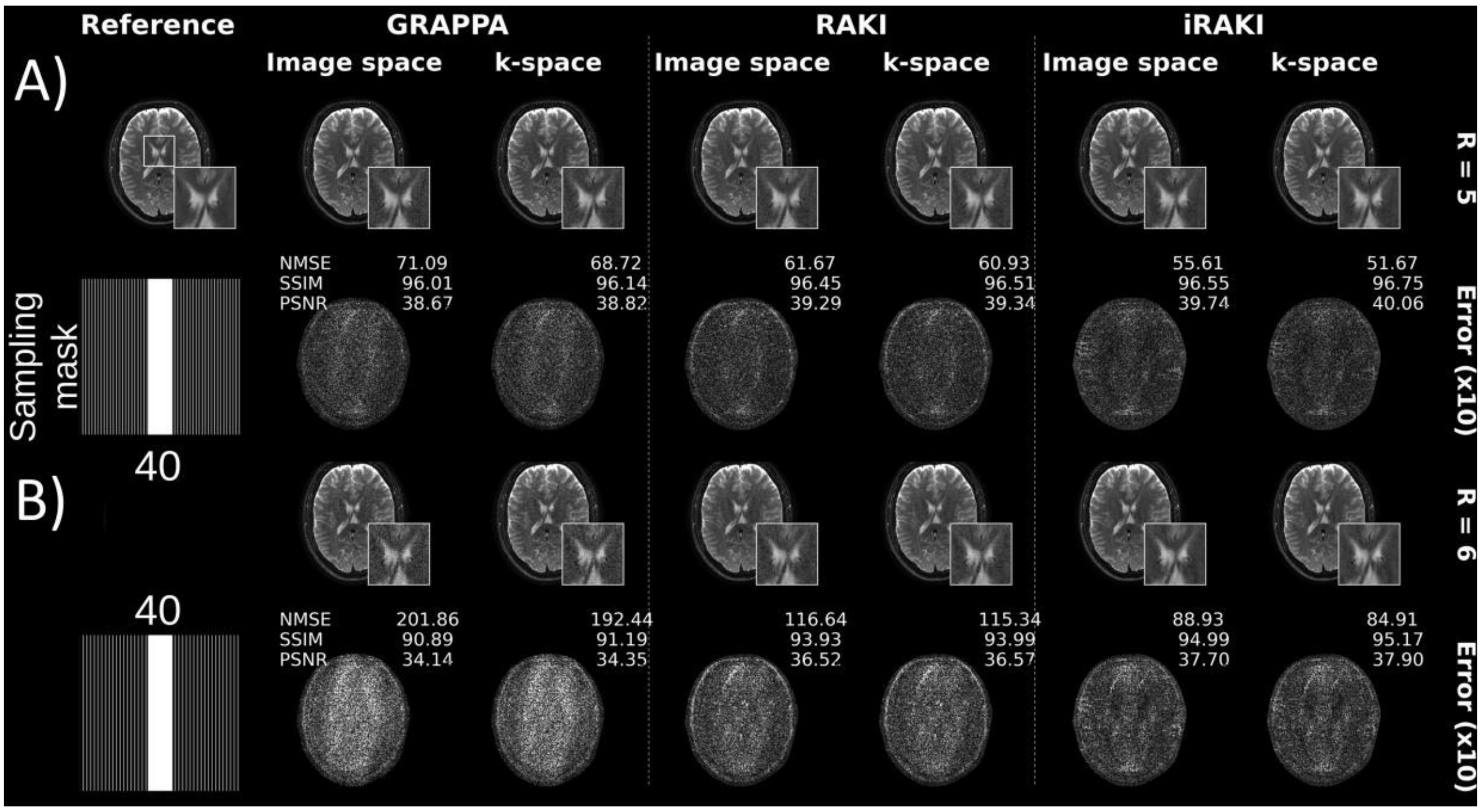

**(A)** GRAPPA, RAKI and iRAKI image reconstructions in k-space (conventional method) and in image space (proposed) for the TSE dataset at $R = 5$ using 40 ACS lines as training data (total acceleration 3.1). Note that the training takes place in k-space, and only the inference step is performed in image space. The error maps are shown below and scaled for display. Quantitative metrics include the normalized mean squared error (NMSE), structural similarity index measure (SSIM), peak signal to noise ratio (PSNR). Both error maps and quantitative metrics show quasi-identical inference in both domains for all reconstructions. $R = 6$ (total acceleration 3.4) is shown in **(B)**. Please see Figure S8 for the imaging scenario with limited training data and enhanced total accelerations.

## Figure S8

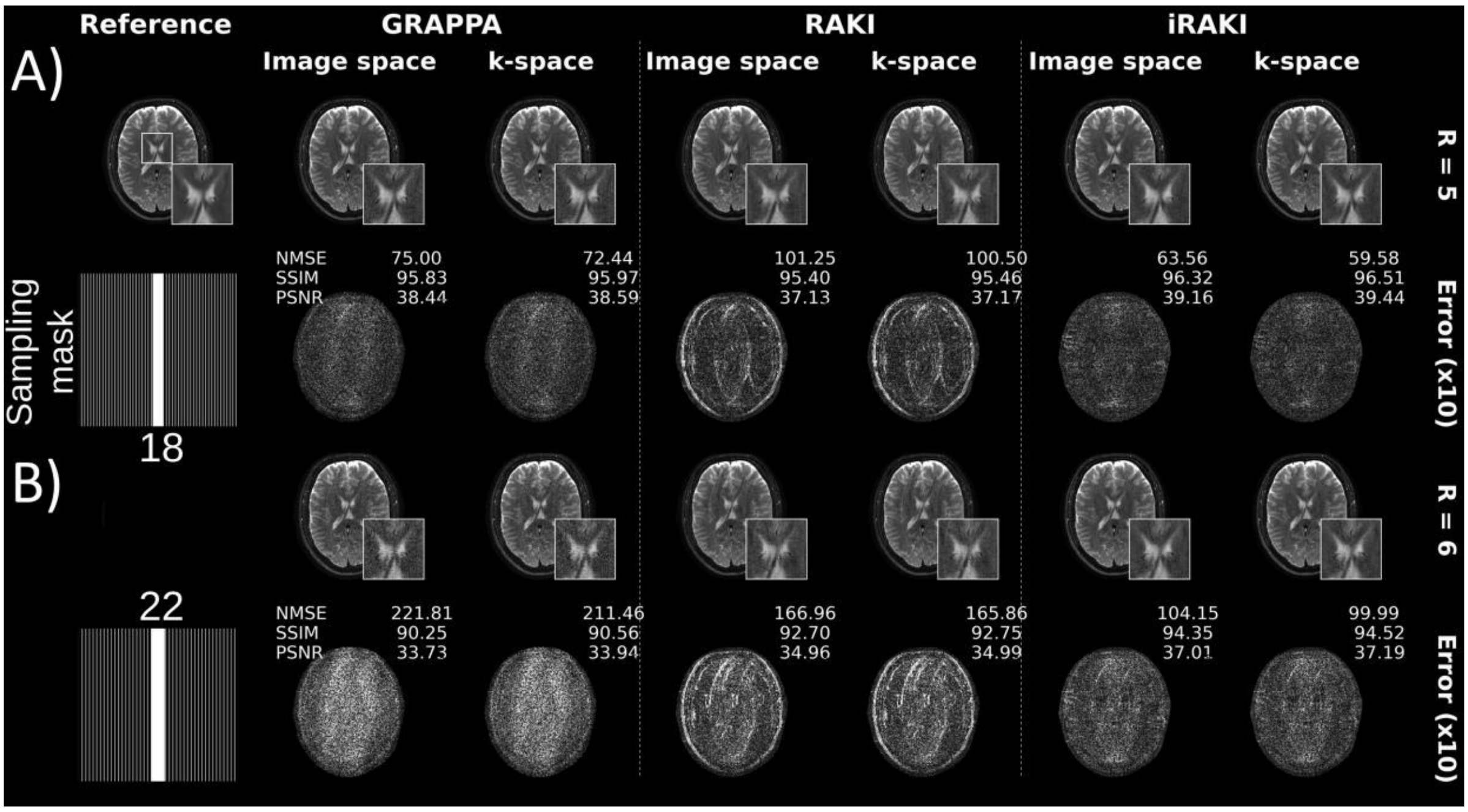

**(A)** GRAPPA, RAKI and iRAKI image reconstructions in k-space (conventional method) and in image space (proposed) for the TSE dataset at $R = 5$ using limited training data amount (only 18 ACS lines, total acceleration 3.9). Residual errors due to the training data limitation are equally displayed in both k-space and image space, supporting the accuracy of the image space formalism. It is worth noting that the iRAKI suppresses residual artifacts in RAKI while providing a similar noise suppression feature. $R = 6$ using only 22 ACS lines (total acceleration 4.2) is shown in **(B)**.

## Figure S9

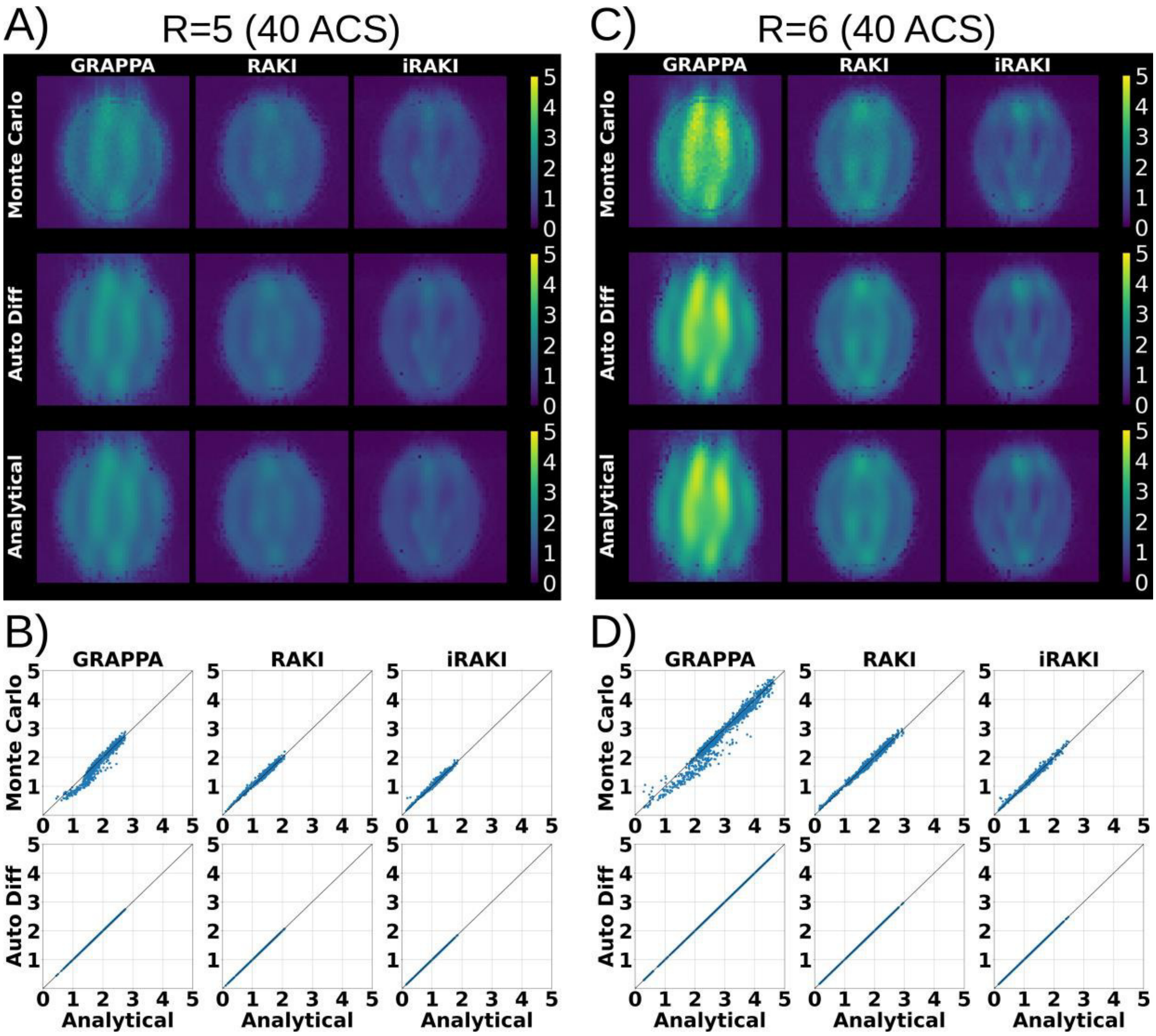

**(A)** G-factor maps (50x50 low resolution) computed via Monte Carlo simulations (1,000 repetitions), via auto differentiation and analytically for GRAPPA, RAKI and iRAKI reconstructions (TSE, $R = 5$, 40 ACS lines). **(B)** G-factors obtained analytically are plotted against those obtained via Monte Carlo (top row) and via auto differentiation (bottom row). The g-factor maps at $R = 6$ and 40 ACS lines are shown in **(C)**, and plots of g-factors obtained analytically against those obtained via Monte Carlo and via auto differentiation are shown in **(D).**

## Figure S10

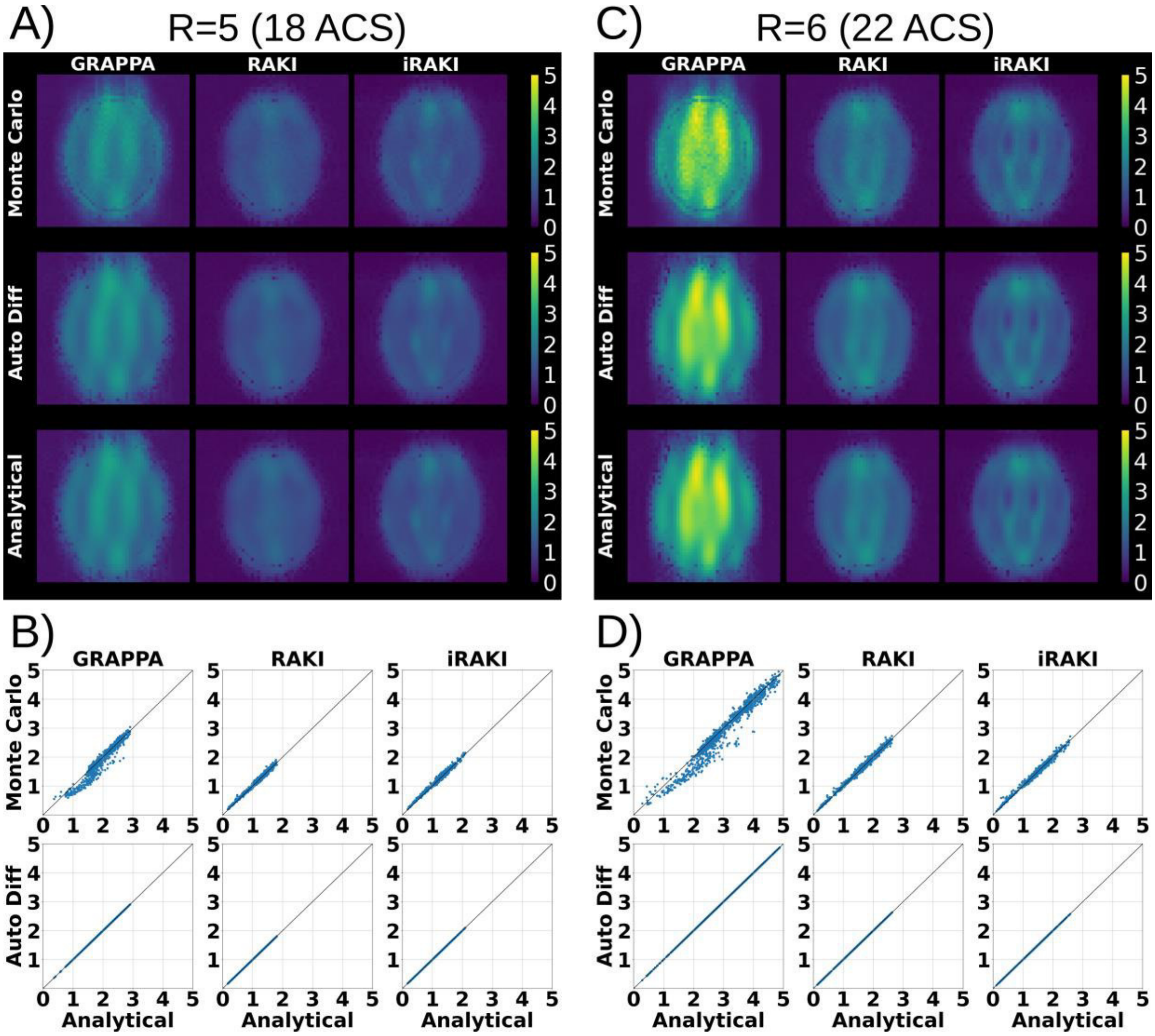

**(A)** G-factor maps (50x50 low resolution) computed via Monte Carlo simulations (1,000 repetitions), via auto differentiation and analytically for GRAPPA, RAKI and iRAKI reconstructions (TSE, $R = 5$, 18 ACS lines). **(B)** G-factors obtained analytically are plotted against those obtained via Monte Carlo (top row) and via auto differentiation (bottom row). The g-factor maps at $R = 6$ and 22 ACS lines are shown in **(C)**, and plots of g-factors obtained analytically against those obtained via Monte Carlo and via auto differentiation are shown in **(D).**

## Figure S11

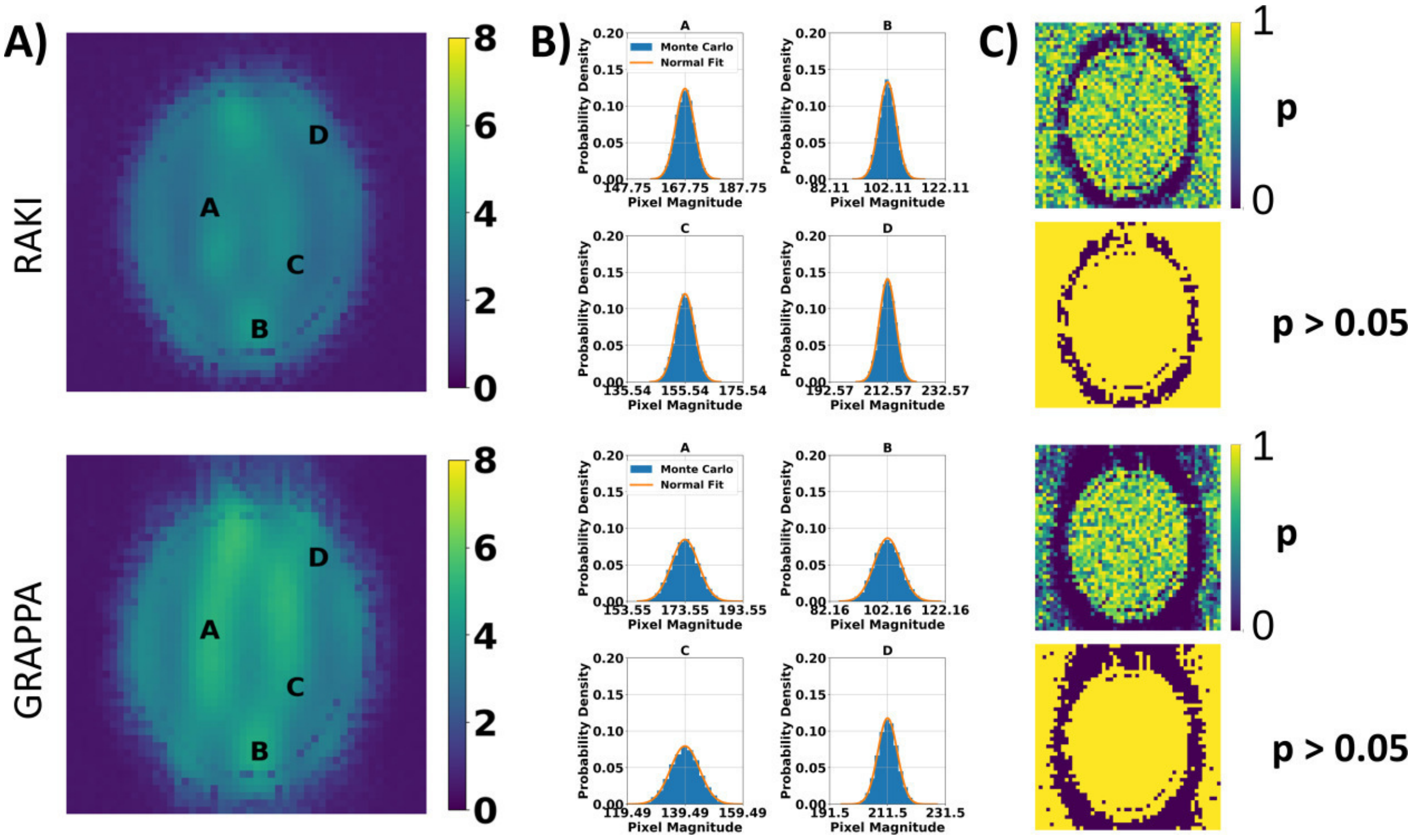

**(A)** Standard deviation maps obtained from Monte-Carlo simulations (10,000 repetitions) for RAKI and GRAPPA reconstructions of the TSE dataset at $R = 5$ (40 ACS lines). **(B)** Voxel magnitude histograms of 10,000 pseudo replicas obtained from voxel locations indexed by A-D in **(A)**, and corresponding fitted normal distributions. **(C)** P-value maps computed in Kolmogorov-Smirnov tests for normality, and binary masks where p>0.05, which is the significance level not to reject the null hypothesis (i.e. voxel magnitude distributions of pseudo replicas are normal). For almost all voxels in the region of interest, a normal distribution can be assumed for both RAKI and GRAPPA, which validates the use of the generalized g-factor computation for RAKI. Please see Figure S12 for $R = 6$.

**Figure S12**

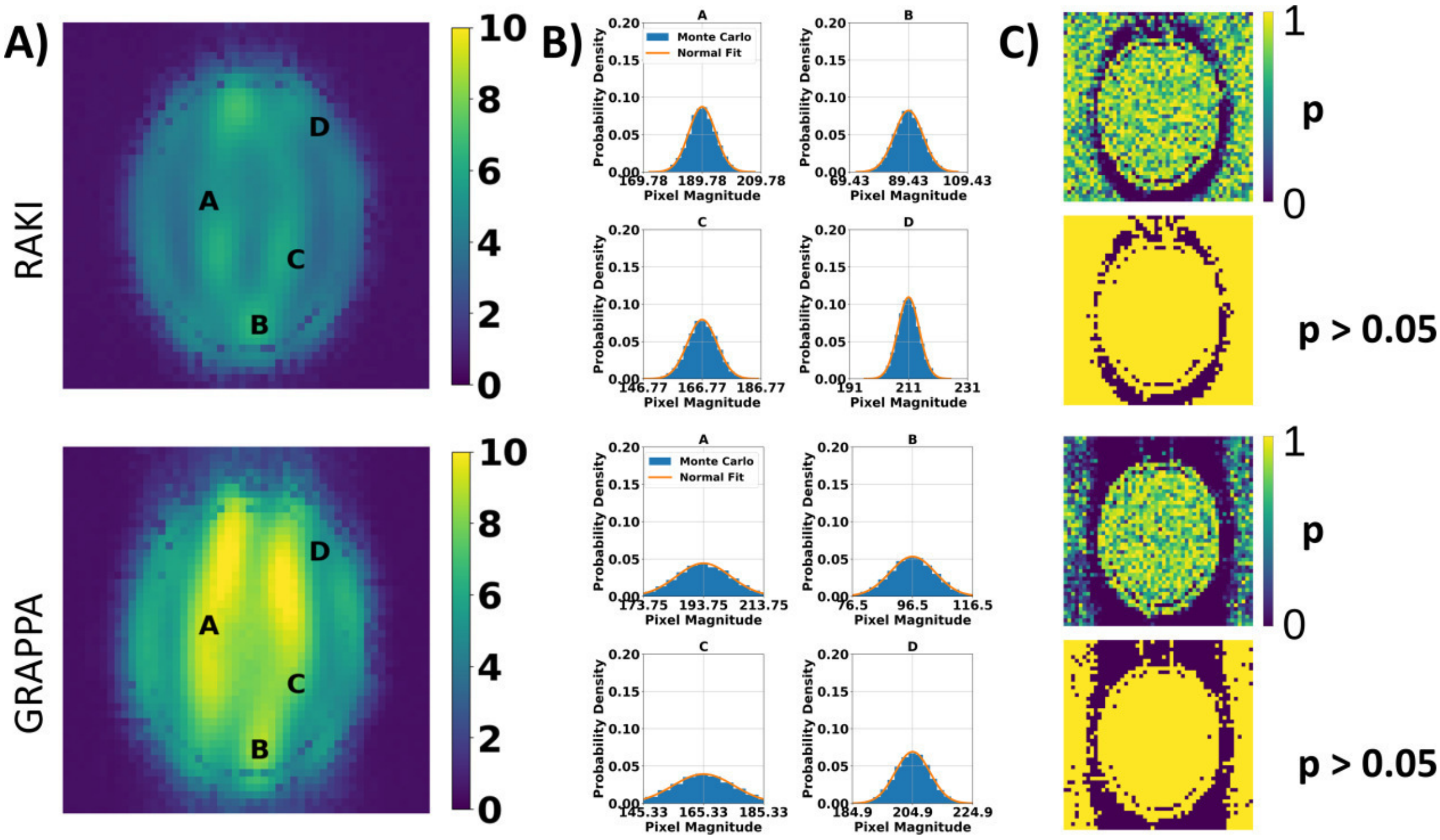

**(A)** Standard deviation maps obtained from Monte-Carlo simulations (10,000 repetitions) for RAKI and GRAPPA reconstructions of the TSE dataset at $R = 6$ (40 ACS lines). **(B)** Voxel magnitude histograms of 10,000 pseudo replicas obtained from voxel locations indexed by A-D in **(A)**, and corresponding fitted normal distributions. **(C)** P-value maps computed in Kolmogorov-Smirnov tests for normality, and binary masks where p>0.05, which is the significance level not to reject the null hypothesis (i.e. voxel magnitude distributions of pseudo replicas are normal). For almost all voxels in the region of interest, a normal distribution can be assumed for both RAKI and GRAPPA, which validates the use of the generalized g-factor computation for RAKI.

## Figure S13

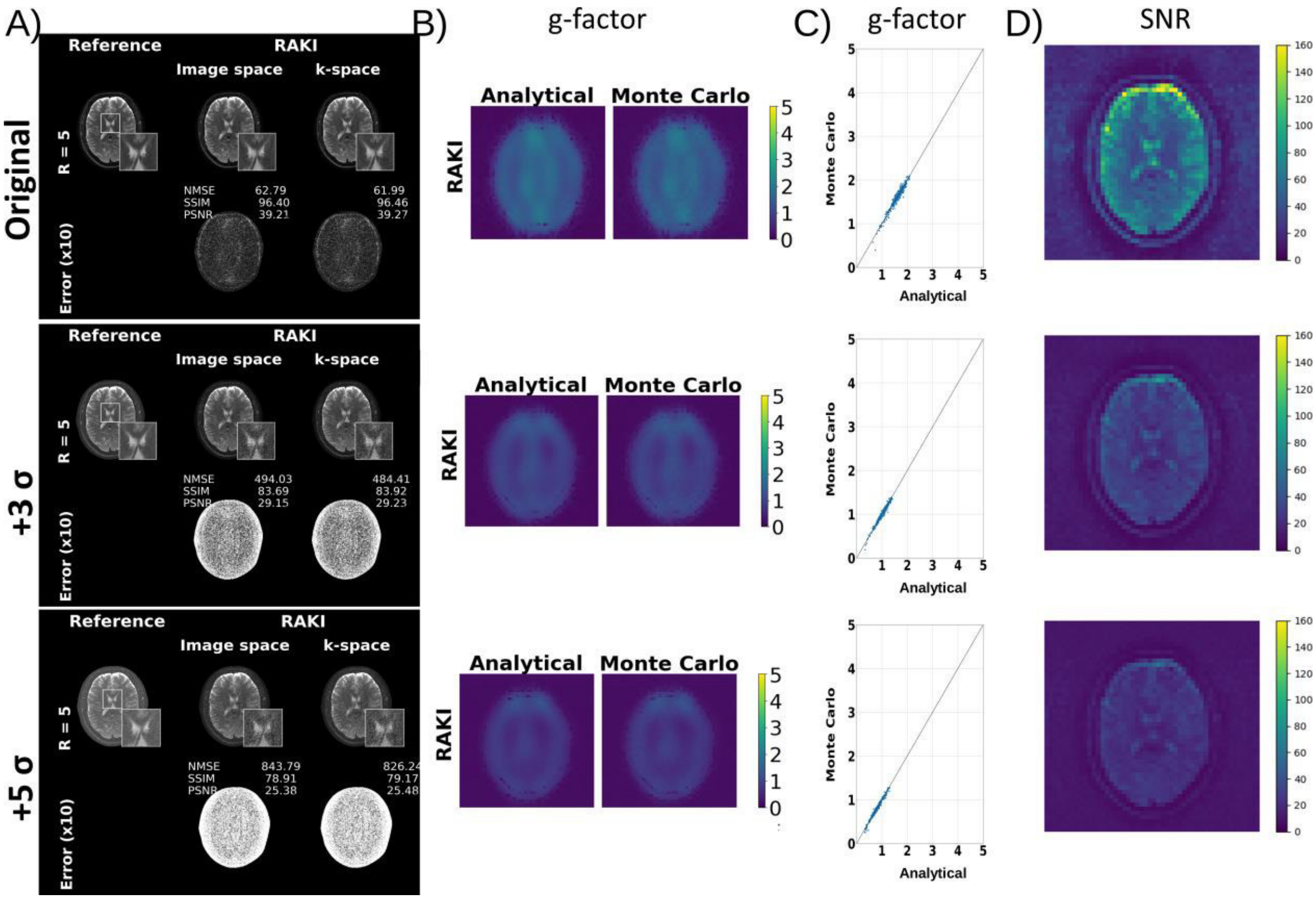

**(A)** RAKI image reconstructions and error maps (scaled for display) for the TSE dataset at $R = 5$ using 40 ACS lines without additive noise superimposition, i.e. original, and with random Gaussian noise superimposition with standard deviation of 3 **σ** and 5 **σ** and zero mean. **(B)** RAKI g-factor maps obtained analytically and via Monte-Carlo simulations (1,000 repetitions). **(C)** G-factors of all voxels computed analytically plotted against g-factors obtained via the Monte Carlo simulations shown in (B). **(D)** SNR maps of images in **(A)** obtained via the Monte-Carlo simulations. Please see Figure S15 for corresponding GRAPPA reconstructions and g-factor maps.

## Figure S14

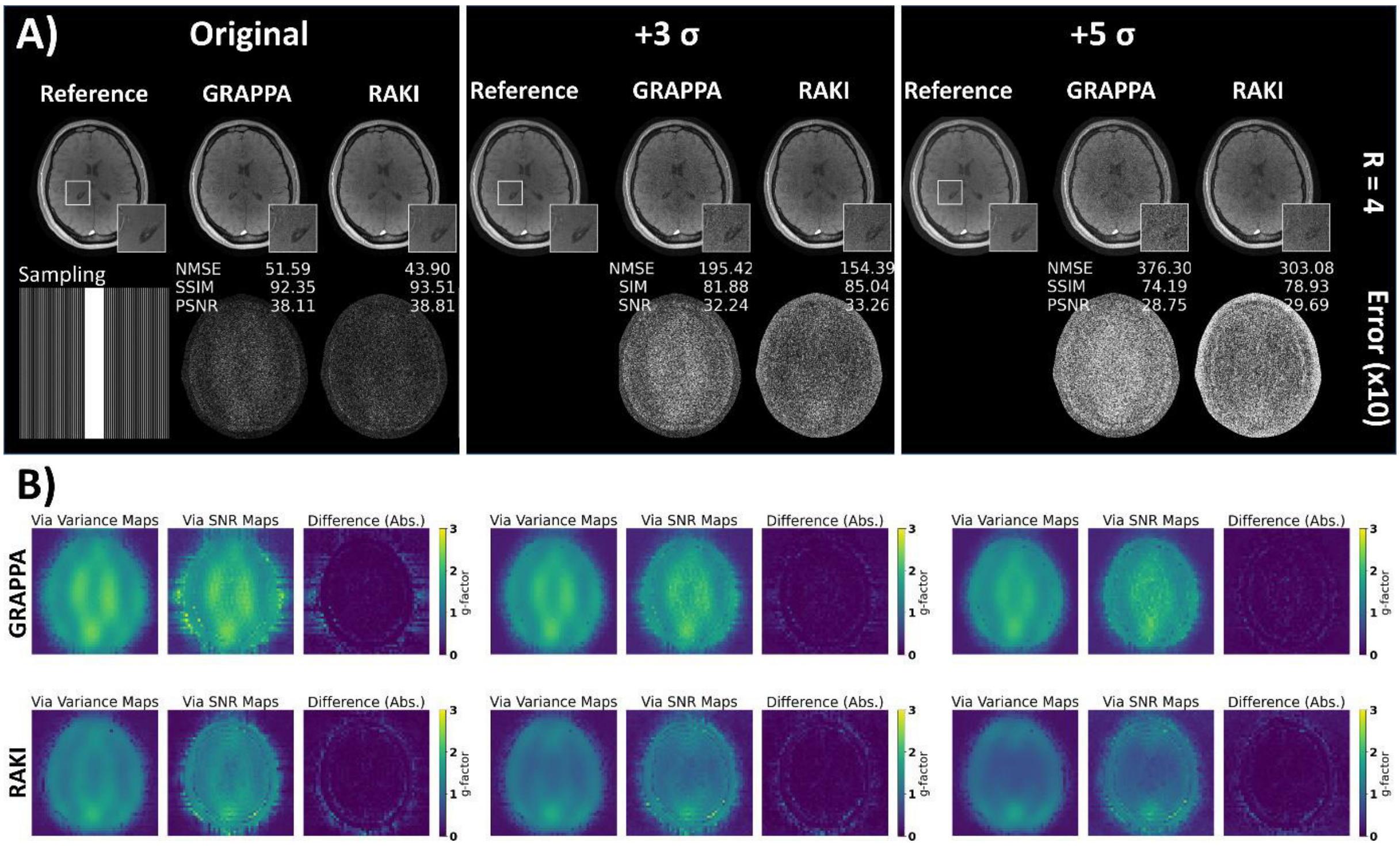

**(A)** Comparison of GRAPPA and RAKI image reconstructions and error maps (scaled for display) for the FLASH dataset at R=4 using 40 ACS lines. The reconstructions are shown for three scenarios: the original dataset without any additive noise, and datasets with added random Gaussian noise (zero mean) at standard deviations of 3σ and 5σ, respectively. The impact of the noise is clearly evident in the error maps. Notably, RAKI consistently demonstrates strong resilience to noise across all conditions, preserving image quality much more effectively than GRAPPA.

**(B)** Comparison of g-factor maps obtained via variance maps versus g-factors obtained via SNR maps (see Eq. (8)). This comparison allows to investigate potential reconstruction biases in RAKI, which may compromise the g-factor calculation via variance and algebraic Jacobians. In an unbiased reconstruction, the mean signal values should match between the reference and the accelerated images, so both methods yield the same g-factor. However, if there would be a reconstruction bias that alters the mean signal, the g-factor calculated from the SNR maps would differ from the one obtained solely from the noise variance. Please note, to yield the variance- and SNR maps, pseudo-replicas (Monte Carlo simulations, 10,000 samples) were computed, and both maps were derived from the same samples in each case.

## Figure S15

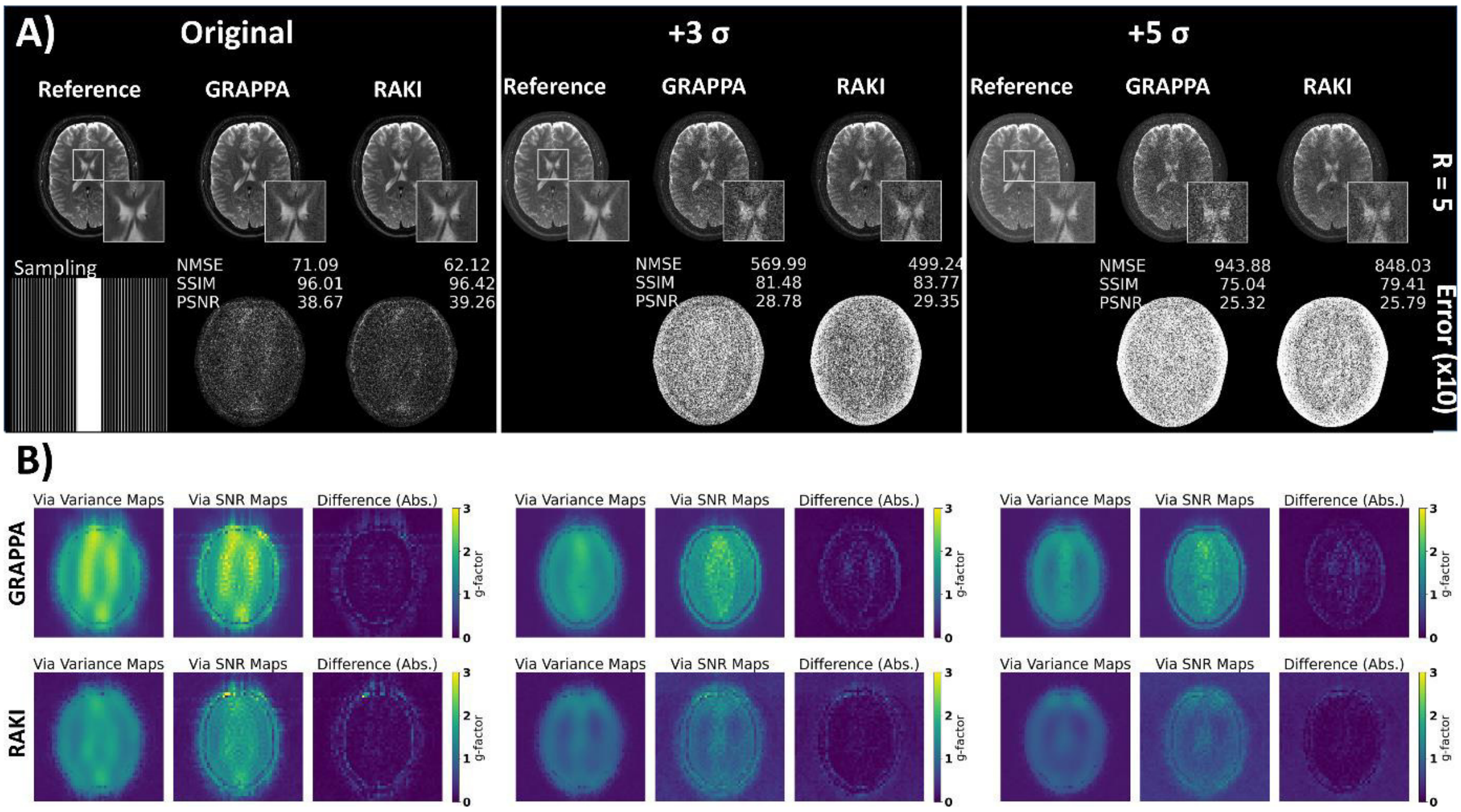

**(A)** Comparison of GRAPPA and RAKI image reconstructions and error maps (scaled for display) for the TSE dataset at R=5 using 40 ACS lines. The reconstructions are shown for three scenarios: the original dataset, and datasets with added random Gaussian noise (zero mean) at standard deviations of 3σ and 5σ, respectively. RAKI demonstrates strong suppression of noise enhancement in all cases in comparison to GRAPPA.

**(B)** Comparison of g-factor maps obtained via variance maps versus g-factors obtained via SNR maps (see Eq. (8)). For both GRAPPA and RAKI, no deviations between both g-factor estimation approaches can be observed. This justifies the RAKI g-factor formulation proposed in this work to estimate the noise propagation in the inference process. Please note, to yield the variance- and SNR maps, pseudo-replicas (Monte Carlo simulations, 10,000 samples) were computed, and both maps were derived from the same samples in each case.

## Supporting Material: Videos (mp4)

### Video S1

Video showing GRAPPA, RAKI and iRAKI reconstructions (FLASH, $R = 4$, 18 ACS lines) in k-space and image space domain for varying negative slope parameter $a$ in ℂReLU activation. Note that the degree of nonlinearity in RAKI and iRAKI can be controlled by adjusting $a$, with $a = 0.0$ representing maximum nonlinearity, and $a = 1.0$ representing a linear model (i.e. GRAPPA with multiple hidden layers). It can be seen that parameter $a$ can serve as regularization parameter in RAKI, which trades-off suppression of noise enhancement against residual artifacts.

### Video S2

Video showing analytical g-factor maps (50x50 low resolution) for GRAPPA, RAKI and iRAKI reconstructions (see Video S1) for varying negative slope parameter $a$ in ℂReLU activation.

### Video S3

Video showing signals to be activated, activation mask and activated signals for the center channel (64.) in the first hidden layer in RAKI for varying negative slope parameter $\alpha$ in ℂReLU activation.

### Video S4

Video showing signals to be activated, activation mask and activated signals for the center channel (64.) in the first hidden layer in iRAKI for varying negative slope parameter $\alpha$ in ℂReLU activation.